\documentclass[10pt,twocolumn,letterpaper]{article}

\usepackage[pagenumbers]{cvpr} % To force page numbers, e.g. for an arXiv version

\usepackage[dvipsnames]{xcolor}

\usepackage{graphicx}
\usepackage{amsmath}
\usepackage{amssymb}
\usepackage{booktabs}
\usepackage{color}
\usepackage{xcolor}
\usepackage{enumerate}
\usepackage{placeins}
\usepackage{lipsum}  
\usepackage{pifont}
\usepackage{amsbsy}
\definecolor{cvprblue}{rgb}{0.21,0.49,0.74}
\usepackage[pagebackref=true,breaklinks=true,colorlinks,bookmarks=false,hypertexnames=false]{hyperref}

\usepackage[capitalize]{cleveref}
\crefname{section}{Sec.}{Secs.}
\Crefname{section}{Section}{Sections}
\Crefname{table}{Table}{Tables}
\crefname{table}{Tab.}{Tabs.}

\def\paperID{208} % *** Enter the CVPR Paper ID here
\def\confName{CVPR}
\def\confYear{2024}

\begin{document}
\def \modelname {\mbox{FlowMDM}}
\def \supp {supp. material} %Appendix
\renewcommand\figureautorefname{Fig.}
\renewcommand\equationautorefname{Eq.}
\renewcommand\tableautorefname{Tab.}
\renewcommand\subsectionautorefname{Sec.}
\renewcommand\sectionautorefname{Sec.}
\newcommand\myeq{\mkern1.5mu{=}\mkern1.5mu}

%\title{MoD: Mixture of Diffusions for \textit{anything}-to-motion generation} % Or HuMoD?? Human Mixture of Diffusions
\title{%FlowMDM: 
%Perpetual Motion Diffusion Model with Blended Positional Encodings}
% for Multi-Control Long-Term Human Motion Generation.}
Seamless Human Motion Composition with Blended Positional Encodings}
% Unleashing the power of relative positional encodings for human motion composition

\author{German Barquero\hspace{0.7cm} Sergio Escalera\hspace{0.7cm} Cristina Palmero \\
Universitat de Barcelona and Computer Vision Center, Spain \\
{\tt\small \{germanbarquero, sescalera\}@ub.edu}, {\tt\small crpalmec7@alumnes.ub.edu} \\
\small\url{https://barquerogerman.github.io/FlowMDM/}
}
%\maketitle
\twocolumn[{%
\renewcommand\twocolumn[1][]{#1}%
\maketitle

\begin{center}
    \centering
    \captionsetup{type=figure}
    \vspace{-2mm}
    \includegraphics[width=\textwidth]{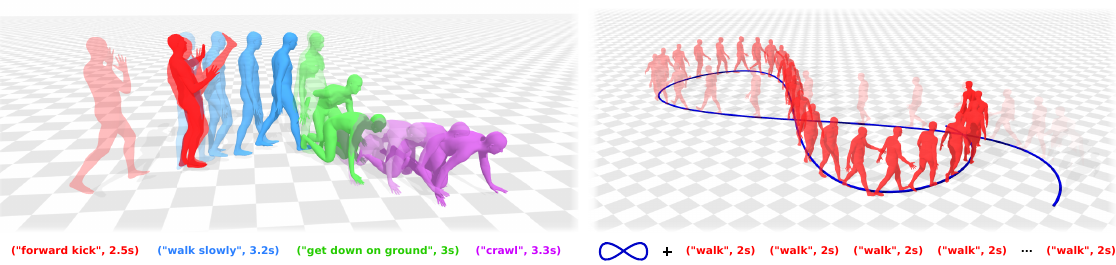}
    \captionof{figure}{We present \modelname{}, a diffusion-based approach capable of generating seamlessly continuous sequences of human motion from textual descriptions (left). The whole sequence is generated simultaneously and it does not require any postprocessing. \modelname{} also makes strides in the challenging problem of extrapolating and controlling periodic motion such as walking, jumping, or waving (right).}
    \label{fig:intro}
\end{center}%
}]

\renewcommand{\arraystretch}{0.75}

%%%%%%%%% ABSTRACT
\begin{abstract}
\vspace{-0.2cm}
Conditional human motion generation is an important topic with many applications in virtual reality, gaming, and robotics. 
While prior works have focused on generating motion guided by text, music, or scenes, these typically result in isolated motions confined to short durations. 
Instead, we address the generation of long, continuous sequences guided by a series of varying textual descriptions. In this context, we introduce FlowMDM, the first diffusion-based model that generates seamless Human Motion Compositions (HMC) without any postprocessing or redundant denoising steps. For this, we introduce the Blended Positional Encodings, a technique that leverages both absolute and relative positional encodings in the denoising chain. More specifically, global motion coherence is recovered at the absolute stage, whereas smooth and realistic transitions are built at the relative stage. As a result, we achieve state-of-the-art results in terms of accuracy, realism, and smoothness on the Babel and HumanML3D datasets. \modelname{} excels when trained with only a single description per motion sequence thanks to its Pose-Centric Cross-ATtention, which makes it robust against varying text descriptions at inference time. Finally, to address the limitations of existing HMC metrics, we propose two new metrics: the Peak Jerk and the Area Under the Jerk, to detect abrupt transitions.
\end{abstract}

%%%%%%%%% BODY TEXT
%\vspace{-0.5cm}
\section{Introduction}
\label{sec:intro}
%\textbf{[1 paragraph] Motivation (seek inspiration in  3rd paragraph of \cite{zhang2023diffcollage}, or \cite{bar2023multidiffusion})}

%\begin{itemize}
%    \item Plenty of shorter data and models for applications A, B, and C, for example. No data for A+B+C. Generating data for A+B+C is important for us.
%    \item Controllable motion generation is of utmost importance for gaming, avatar creation, etc. Being able to generate motion from several inputs such as natural language, target objects, other interactants, etc, is important. However, there is no single solution for all of them.
%    \item Efficiency for such methods is important, and we can't often generate data auto-regressively, as it is often slow, and tends to degenerate after errors accumulation.
%    
%\end{itemize}

In the field of computer vision, recent progress has been made in developing photorealistic avatars~\cite{ma2021pixel} for applications like virtual reality, gaming, and robotics~\cite{park2022metaverse, suzuki2022augmented}. Aside from looking visually realistic, these avatars must also move in a convincing manner.
%While achieving a realistic visual appearance is crucial, ensuring that these avatars move convincingly is equally challenging. 
This is challenging due to the intricate nature of human motion, strongly influenced by various factors such as the environment, interactions, and physical contact~\cite{borges2013video}. Furthermore, complexity increases when attempting to control these motions.
%Recent advancements include the creation of datasets that pair motion data with textual descriptions~\cite{}, and the development of methods to generate motion sequences based on textual descriptions~\cite{}. 
Recent advances include the generation of motion sequences from \textit{control signals} like textual descriptions or actions~\cite{zhu2023human}; however, such methods only produce isolated, standalone motion. Therefore, these approaches fail to handle scenarios where a long motion is driven by distinct control signals on different time slices. Such capability is needed to provide full control over the sequence of desired actions and their duration.
In these scenarios, the generated motion needs to feature seamless and realistic transitions between actions.
% I am lacking here a definition for HMC, and then mentioning its importance.
In this work, we tackle this problem, which we refer to as generative Human Motion Composition (HMC). In particular, we focus on generating single-human motion from text, illustrated in \autoref{fig:intro}. %Nonetheless, the techniques presented in this work can be applied to any type of control signal such as the scene, objects, or even music. %, is of paramount importance in gaming, virtual reality, and robotics.

%\textbf{[2 paragraphs] How the state of the art approaches this.}

One of the primary obstacles in HMC is the lack of datasets that offer long motion sequences with diverse textual annotations. Existing datasets typically feature sequences of limited duration, often lasting only up to 10 seconds, and with just a single control signal governing the entire sequence~\cite{plappert2016kit,guo2022humanml3d}. This limitation calls for innovative solutions to address the inherent complexities of the task. 
Prior works have tackled this problem mostly with autoregressive approaches~\cite{zhang2020perpetual, li2023sequential, athanasiou2022teach, qian2023breaking, lee2023multiact}.
These methods iteratively create compositions by using the current motion as a basis to generate subsequent motions. However, they require datasets with multiple consecutive annotated motions, and tend to degenerate in very long HMC scenarios due to error accumulation~\cite{li2017auto}.
%Despite their good performance due to its convenient inductive bias for motion generation, they tend to degenerate in very long HMC scenarios due to error accumulation~\cite{li2017auto}.
Other recent works have leveraged the infilling capabilities of motion diffusion models to generate motion compositions~\cite{zhang2023diffcollage, shafir2023priormdm}. However, for these, a substantial portion of each motion sequence is generated independently from adjacent motions, and generating transitions requires computing redundant denoising steps. In this work, we propose a novel architecture designed to address these specific challenges. Our main contributions are:

%To address these limitations, we propose \modelname{}. , a model able to generate sequences of arbitrary length controlled by a set of consecutive textual descriptions.
%In particular, our main contributions are:
\begin{itemize}
    %\item We propose \modelname{}, a novel model for human motion composition. Ours is the first non-autoregressive model that generates composed motion in a single inference step without any postprocessing. \modelname{} is able to learn to generate transitions among different actions without explicit supervision. This is possible thanks to 1) blended positional encodings, which makes the learnt motion prior translation-invariant, and 2) the pose-centric cross-attention, which reduces out-of-distribution representations around unseen transitions. As a result, \modelname{} can generate an arbitrary number of condition-controlled motion segments in parallel, all with valid and realistic transitions between them. 
    
    \item We propose \modelname{}%\footnote{Code and pretrained weights are publicly available in **BLINDED**}
    , the first diffusion-based model that generates seamless human motion compositions without any postprocessing or extra denoising steps. To accomplish it, we introduce Blended Positional Encodings (BPE), a new technique for diffusion Transformers that combines the benefits of both absolute and relative positional encodings during sampling. In particular, the denoising first exploits absolute information to recover the global motion coherence, and then leverages relative positions to build smooth and realistic transitions between actions.
    As a result, \modelname{} achieves state-of-the-art results in terms of accuracy, realism, and smoothness in the HumanML3D~\cite{guo2022humanml3d} and Babel~{\cite{punnakkal2021babel}} datasets.
    
    \item We introduce a new attention technique tailored for HMC: the Pose-Centric Cross-ATtention (PCCAT). This layer ensures each pose is denoised based on its own condition and its neighboring poses. Consequently, \modelname{} can be trained on a dataset with only a single condition available per motion sequence and still generate realistic transitions when using multiple conditions at inference time. 
    
    %More specifically,, the learnt human motion prior becomes robust against conditioning shifts, even though these were never seen at training.
    
    %\item We explore, for the first time, the usage of relative embeddings in Transformers for diffusion-based human motion generation. We show their promising capabilities in terms of human motion composition and extrapolation.
    
    \item We reveal the lack of sensitivity of current HMC metrics to identify discontinuous or sharp transitions, and introduce two new metrics that help to detect them: the Peak Jerk (PJ) and the Area Under the Jerk (AUJ).
\end{itemize}

\section{Related work}
\label{sec:relatedwork}

\textbf{Conditional human motion generation.}
Recent studies in motion generation have shown notable progress in synthesizing movements conditioned on diverse modalities such as text~\cite{petrovich2022temos, zhang2022motiondiffuse, dabral2023mofusion, tevet2022mdm, yuan2023physdiff, kim2023flame, guo2022humanml3d, tevet2022motionclip, jiang2023motiongpt, guo2022tm2t, zhang2023t2m}, music~\cite{tseng2023edge, li2021dance, sun2020deepdance, alexanderson2023dance, zhuang2022music2dance, ye2020choreonet, chen2021choreomaster}, scenes~\cite{yi2023scene, cao2020scene, wang2021scene, wang2022scene, hassan2021scene, wang2021scene2}, interactive objects~\cite{xu2023interdiff, adeli2021tripod, corona2020context, kulkarni2023nifty}, and even other humans' behavior~\cite{xu2022stochasticmulti, barquero2022didn, barquero2022comparison, guo2022multi, tanke2023social}. Traditionally, these approaches have been designed to generate motion sequences matching a single condition. The progress of this domain has been boosted by the release of big datasets including diverse modalities or manual annotations~\cite{lin2023motion, guo2022humanml3d, punnakkal2021babel, plappert2016kit, li2021dance, palmero2022chalearn, guo2022multi, bhatnagar2022behave}.
Research has also focused on problems like human motion prediction~\cite{yuan2020dlow, tian2023transfusion, sun2023towards, mao2021gsps, walker2017theposeknows, aliakbarian2021contextually, salzmann2022motron, ma2022multiobjective} and motion infilling~\cite{li2022inbetween, kim2022inbetween, qin2022inbetween, ren2023inbetween, zhou2020inbetween, harvey2020inbetween, starke2023inbetween, kaufmann2020inbetween, oreshkin2023inbetween, li2023example}, which do not rely on extensive manual annotations but rather on motion itself.
Both tasks share a common challenge with HMC: the synthesized motion must not only be plausible but also integrate seamlessly with the neighboring behavior, ensuring fluidity and continuity. In this context, the utilization of human motion priors has been proven to be a successful technique to ensure any generated motion includes natural transitions~\cite{barquero2023belfusion, li2021task, xu2021exploring}. In line with these approaches, our method learns a motion prior specifically tailored for HMC.
%We argue that, given the limitations in motion length and annotations variety of the available datasets, motion priors are a powerful tool for the generation of arbitrarily long motion.

% ===========================================

\textbf{Autoregressive human motion composition.}
%Generating realistic transitions is also of utmost importance for generating long motion sequences featuring different actions. 
%For images, an equivalent problem would be large-size image generation, such as panoramic images or images with atypical aspect ratios. Prior work addressing these problems have presented approaches where the large image is decomposed into overlapping patches~\cite{bar2023multidiffusion, zhang2023diffcollage}. 
As in many other sequence modeling tasks, HMC was also first tackled with autoregressive methods. The gold standard has been pairing variational autoencoders with autoregressive decoders such as recurrent neural networks~\cite{zhang2020perpetual} or Transformers~\cite{li2023sequential, athanasiou2022teach, qian2023breaking, lee2023multiact}. Alternative approaches have introduced specialized reinforcement learning frameworks~\cite{zhang2022wanderings, luo2023perpetual, yang2023synthesizing}. % to ensure that the motion satisfies real-world constraints
Autoregressive models rely on the availability of annotated motion transitions, a requirement that constrains the robustness of the models due to the scarcity of such data. To mitigate this issue, some methods include additional postprocessing steps like linear interpolations~\cite{athanasiou2022teach}, or affine transformations~\cite{lee2023multiact}. However, these can distort the human motion dynamics and require a predetermined estimation of the transitions duration. Furthermore, autoregressive approaches generate motion solely based on the preceding motion. We argue that an accurate model should mimic the humans innate capacity to anticipate their next action and adapt their current behavior accordingly~\cite{kunde2007no, engstrom1996reaction}.

\textbf{Diffusion-based human motion composition.}
Diffusion models have excelled at conditional generation~\cite{sohl2015deep, croitoru2023diffusion, ho2020denoising}. 
They also possess great zero-shot capabilities for image inpainting~\cite{rombach2022high}, and its equivalence in motion: motion infilling. DiffCollage~\cite{zhang2023diffcollage}, MultiDiffusion~\cite{bar2023multidiffusion}, and DoubleTake~\cite{shafir2023priormdm} proposed to modify the diffusion sampling process to simultaneously generate temporally superimposed motion sequences, and combine the estimated noise in the overlapped regions so that an infilled transition emerges. DoubleTake complemented such overlapped sampling with a refinement step in which the emerged transition undergoes further unconditional denoising steps. All these methods share two main limitations.
First, they are constrained to modeling dependencies among neighboring motion sequences. This becomes a limitation when three or more consecutive actions share semantics and collectively represent a more comprehensive action. In this case, the motion dependencies may extend beyond contiguous actions.
Second, they need to set the number of frames that each transition takes between consecutive actions, for which extra computations are incorporated.
%Second, they incorporate extra computations for the overlapped regions, and require to set the duration of each transition between consecutive actions.
Our work seeks to address these constraints by offering a solution able to model longer inter-sequence dynamics without imposing extra computational burdens or predefined transition durations.

\def\subseq{\mathcal{S}}
\def\consecsubsec{\subseq_{i}\rightarrow \subseq_{i+1}}
\def\consecsubsecABC{\subseq_{i}\rightarrow \subseq_{i+1}\rightarrow \subseq_{i+2}}
\def\poseA{x_{m}}
\def\poseB{x_{n}}
\def\cond{c}
\def\condA{\cond_{m}}
\def\condB{\cond_{n}}
\def\embA{E_{\poseA,\condA}}
\def\embB{E_{\poseB,\condB}}
\def\embBnocond{E_{\poseB}}

\section{Methodology}
\label{sec:methodology}
%TO BE DEFINED FOR CONSISTENCY: subsequence, control signal?

%\begin{itemize}
%    \item \textbf{Q. Is this effect architecture dependent? Can it work with autoencoding denoising models?} I think the structure needs to be non-autoencoding (not working for U-Nets). Otherwise the intermediate representation might be too specific to the target field. We'll need to check with MDM + Encoder configuration. Key concept here: the network needs to be \textbf{translation invariant}. Denoised pose at timestep $t_1$ is the same as in timestep $t_2$ given the same contextual variables.
%\end{itemize}

%\subsection{Problem definition}

%\textbf{[1 paragraph]} Defining our problem. Training with homogeneously conditioned sequences, and we want to infer arbitrarily long motion sequences that are heteregeneously conditioned.

\textbf{Problem definition.} Our goal consists in generating a motion sequence %$\mathcal{M}$ 
of $N$ frames, with the capability of conditioning the generated motion inside non-overlapping intervals $[0, \tau_1), [\tau_1, \tau_2),..., [\tau_{j}, N)$, with $0 {<} \tau_1 {<} \cdots {<} \tau_j {<} N$.
%and $\tau_{i+1}{-}\tau_{i}{<} L$ for any $i$. 
We will refer to the motion inside these intervals as \textit{motion subsequences}, or $\subseq_{i}=\lbrace x_{\tau_i}, ..., x_{\tau_{i+1}-1} \rbrace$, each driven by its corresponding condition $c_i$, and with a maximum length of $L$. %, with $c$ being the corresponding control signal.
It is essential that consecutive subsequences, influenced by different control signals, transition seamlessly and realistically. In particular, we aim at the even more challenging case where motion sequences containing several pairs of $(S_i, c_i)$ are not necessarily available in our dataset.

%In particular, we encounter three issues:
%\begin{enumerate}
%\item \textcolor{red}{[Encoder-only diffusion]} 
%\item \textcolor{red}{[Heterogeneous conditioning]}
%Lack of datasets with arbitrarily long sequences, with diverse control signals along the time axis.
%Prior methods can not deal with an heterogeneous set of control signals.
%\item \textcolor{red}{[PE]} Human motion datasets are limited in size and diversity. Therefore, transitions from any A to any B are generally not available. To overcome the data limitation, our model should learn a robust \textit{human motion prior} able to understand the kinematics of the human body.
%\end{enumerate}

% [SUMMARY OF METHOD]
In this section, we present \modelname{}, an architecture with strong inductive biases that promote the emergence of a \textbf{robust translation-invariant motion prior}. Such \textit{motion prior} is learned with a diffusion model equipped with a bidirectional (i.e., encoder-only) Transformer, similar to prior works~\cite{tevet2022mdm, shafir2023priormdm}. With it, we overcome the main limitations of autoregressive methods (\autoref{subsec:method_encoder_only}). However, previous works are constrained in terms of motion duration. We could arguably provide extrapolation capabilities to the diffusion model by replacing the absolute positional encoding with a relative alternative, thus making the denoising of each pose \textit{translation invariant}. However, this technique would fail to build complex compositional semantics that require knowledge about the start and end of each subsequence. 
For example, when generating the motion composition $S_{i}{\rightarrow} S_{i+1}$ with $c_{i}{=}\textit{`walking'}$ and $c_{i+1}{=}\textit{`walk and sit down'}$, $S_{i+1}$ might only feature the action `\textit{sit down}' because, with only relative positional information, the Transformer cannot know if the partially denoised `\textit{walking}' motion preceding the beginning of $S_{i+1}$ belongs to $S_{i}$ or $S_{i+1}$. 
%For example, when denoising (i.e., generating) the motion composition $S_{i}\rightarrow S_{i+1}$ with $c_{i}{=}\textit{`walking'}\rightarrow c_{i+1}{=}\textit{`walk and sit down'}$, $S_{i+1}$ might only feature the action `\textit{sit down}' because, from a relative point of view, $S_{i+1}$ is complete, as it misinterprets that the preceding `\textit{walking}' motion from $S_{i}$ motion belongs to $S_{i+1}$. 
To combine the benefits of both relative and absolute positional encodings, we introduce BPE (\autoref{subsec:method_blendedpe}). This novel technique exploits the iterative nature of diffusion models to promote intra-subsequence global coherence in earlier denoising stages, while making later denoising stages translation invariant, ensuring that realistic and plausible transitions naturally emerge between subsequences. Still, during training, the condition remains unchanged throughout all ground truth motion sequences.
In order to make our denoising model \textit{robust} to having multiple conditions per sequence at inference, we introduce a new attention paradigm called PCCAT (\autoref{subsec:method_conditioning}). As a result, \modelname{} is able to simultaneously generate very long compositions of human motion subsequences, all in harmony and fostering plausible transitions between them, without explicit supervision on transitions generation.

\subsection{Bidirectional diffusion}
\label{subsec:method_encoder_only} %In fact, humans constantly anticipate their next action and adapt their ongoing motion so that transitioning becomes feasible given their body dynamics and movement ranges.
The cumulative nature of errors in autoregressive models often results in a decline in performance when generating long sequences~\cite{li2017auto}. This is exacerbated in HMC, where transitions are scarce or even missing in the training corpus, and the model needs to deal with domain shifts at inference.
Another limitation of autoregressive methods is that the generated $\subseq_{i}$ only depends on $\lbrace \subseq_{j}\rbrace_{j<i}$. We discussed in \autoref{sec:relatedwork} why this is a suboptimal solution for HMC.
Thus, an appropriate model for HMC should also be able to anticipate the following motion, $\subseq_{i+1}$, and possibly adapt $\subseq_i$ so that the transition is feasible. 
We argue that the iterative paradigm of diffusion models provides very appropriate inductive biases for naturally mimicking such ability: the partially denoised $\subseq_i$ and $\subseq_{i+1}$ are refined later in successive denoising steps. By choosing a bidirectional Transformer as our denoising function~\cite{devlin2018bert}, we enable the modeling of both past and future dependencies. Therefore, we design our framework as a bidirectional motion diffusion model, similar to MDM~\cite{tevet2022mdm}. We refer the reader to \cite{yang2022diffusion} for more details on the theoretical aspects of diffusion models.
% (\textcolor{red}{T refers to Markov's chain??}):

%\begin{equation}
%\label{eq:loss}
%\begin{gathered}
%    \mathcal{L}(X, C)= 
%    \sum^T_{t=1}
%    \underset{q(x_t|x_0)}{\mathbb{E}}
%    \|f_{\phi}(x_t, t, C)) - x_0)\|_{2}.
%    \vspace{-0.1cm}
%\end{gathered}
%\end{equation}

%In order to mix motion generation models, all of them must be \textbf{bidirectional}, or encoder-only, models. \textbf{Or cross-attention ones, similar to a language-translation where original space is noisy, and target space is denoised.} This is of utmost importance so that denoising in transitions incorporates information from both the previous and following denoised motion. This is shown in \autoref{tab:ablation_bidirectional}. Although $M_2$ might not see motions starting from sitting position, $M_1$ knows how to transition from sitting position to standing position, so that common denoising will make this transition emerge (for the sake of motion smoothness and dynamics).

\subsection{Blended positional encodings}
\label{subsec:method_blendedpe}

Diffusion models can learn strong motion priors that ensure any motion generated is realistic and plausible~\cite{shafir2023priormdm}. In fact, they can also generate smooth transitions between subsequences~\cite{shafir2023priormdm, zhang2023diffcollage, bar2023multidiffusion}. However, these capabilities stem from inference-time motion infilling techniques, which we argue do not exploit the full potential of human motion priors. In fact, building a prior that extrapolates well to sequences longer than those observed during training is very challenging. The field of natural language processing has made progress in sequence extrapolation techniques, notably by substituting absolute positional encoding (APE) with a relative (RPE) counterpart~\cite{kazemnejad2023impactrpe}. By only providing information regarding how far tokens are between them, they achieve sequence-wise translation invariance and, therefore, can extrapolate their modeling capabilities to longer sequences. %Yet, for composite motion actions, the poses order alone is insufficient. 
%One of our hypotheses is that diffusion models can learn strong motion priors so that any motion generated is realistic and plausible. If this holds, they could also generate a realistic transition between $\subseq_{i}$ and $\subseq_{i+1}$. Nevertheless, such capability would remain bounded by the longest motion observed during training, making the generation of very long sequences impossible. The field of natural language processing has made progress in sequence extrapolation techniques, notably by substituting absolute positional encoding (APE) with a relative (RPE) counterpart~\cite{}. By only providing information regarding how far tokens are between them, they achieve translation invariance within the sequence and, therefore, can extrapolate their modeling capabilities to longer sequences~\cite{}. %Yet, for composite motion actions, the poses order alone is insufficient. 
Yet, the absolute positions of poses within a motion, including their distances to the start and end of the action, are necessary to build the global semantics of the motion, as exemplified at the beginning of this section.

\begin{figure}

    \centering
    \includegraphics[width=\linewidth]{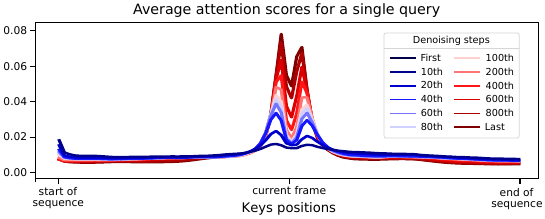}
    \vspace{-7mm}
    \caption{Attention scores of a single query pose (current frame) as a function of the pose attended to (x-axis) in a diffusion-based motion generation model with a sinusoidal absolute positional encoding. Curves show the scores at each denoising step. We observe that, whereas early steps show strong global dependencies (blue), later denoising stages exhibit a clearly local behavior (red).}
    \label{fig:global_vs_local_att}
    \vspace{-3mm}

\end{figure}
Here, we propose BPE, a novel positional encoding scheme designed for diffusion models that enables motion extrapolation while preserving the global motion semantics. Our BPE is inspired by the observation that in motion, high frequencies encompass local fine details, whereas low frequencies capture global structures. Similar insights have been drawn for images~\cite{pan2022fast}. Diffusion models excel at decomposing the generation process into recovering lower frequencies, and gradually transitioning to higher frequencies. \autoref{fig:global_vs_local_att} shows how at early denoising phases, motion diffusion models prioritize global inter-frame dependencies, shifting towards local relative dependencies as the process unfolds.
The proposed BPE harmonizes these dynamics \textit{during inference}: at early denoising stages, our denoising model is fed with an APE and, towards the conclusion, with an RPE. A scheduler guides this transition.
As a result, intra-subsequence global dependencies are recovered at the beginning of the denoising, and intra- and inter-subsequences motion smoothness and realism are promoted later.
To make the model understand APE and RPE at inference, we expose it to both encodings by randomly alternating them during training. As a result, the BPE schedule can be tuned at inference time to balance the intra-subsequence coherence and the inter-subsequence realism trade-off.
%To make this possible, we randomly alternate between APE and RPE during training. This way, the model learns to understand both encodings.
%\textcolor{purple}{Yet,  the disparate feeding mechanisms of absolute and relative positional encodings at various model stages can introduce severe domain shifts in intermediate representations, potentially hindering learning.} To alleviate it, we propose to build both RPE and APE under the same paradigm: rotary position embeddings~\cite{su2021roformer}.

%As a result, the motion diffusion model learns to exploit the positional information available at that time. Then, at inference time, we can define a schedule $h(t)$ that selects  the appropriate APE or RPE depending on the needs of the denoising step. In particular, $h(t)$ provides an absolute positional embedding when $t\rightarrow T$ and transitions to providing relative embeddings when $t\rightarrow 0$.

\textbf{Rotary Position Encoding (RoPE).} Our choice for RPE is rotary embeddings~\cite{su2021roformer}. RoPE integrates a position embedding into the queries and keys, ensuring that after dot-product multiplication, the attention scores' positional information reflects only the relative pairwise distance between queries and keys. Specifically, let $W_q$ and $W_k$ be the projection matrices into the $d$-dimensional spaces of queries and keys. Then, RoPE encodes the absolute positions $m$ and $n$ of a pair of query ($q_m{=}W_q x_m$) and key ($k_n{=}W_k x_n$), respectively, as $d$-dimensional rotations $R^d_{m}, R^d_{n}$ over the projected poses $x_m, x_n$. The rotation angles are parameterized by $m$ and $n$ so that the attention formulation becomes:
\begin{equation}\small
q^T_mk_n=(R^d_m W_q x_m)^T(R^d_n W_k x_n) = x_m^T W_q R^d_{n-m} W_k x_n.
\label{eq:relative_rope}
\end{equation}

%We also include 

%\begin{equation}
%\langle f_q(x_m, m), f_k(x_n, n)\rangle = g(x_m, x_n, m-n).
%\label{eq:relative_rope}
%\end{equation}

Note that the resulting rotation $R^d_{n-m}$ only depends on the distance between $n$ and $m$, and any absolute information about $n$ or $m$ is removed. RoPE is a natural choice for our RPE due to its simplicity and convenient injection before the attention takes place. As a result, RoPE is compatible with faster attention techniques like FlashAttention~\cite{dao2022flashattention, dao2023flashattention2}.

\textbf{Sinusoidal Position Encoding.} Our APE is the classic sinusoidal position encoding~\cite{vaswani2017attention}, which leverages sine and cosine functions to inject positional information. It is added to the queries, keys, and values of the attention layers.

Note that for APE, attention is limited to each subsequence, while for RPE, attention spans all frames up to the attention horizon $H{<}L{<}N$. Since $L$ defines the maximum range of motion dynamics learned during RPE training, there is no advantage in setting $H{\geq}L$ (Tabs. D/E in supp. material). Leveraging both APE and RPE constraints ensures quadratic complexity over the maximum subsequence length $L$ in both memory and computation~\cite{beltagy2020longformer}. 
%Linear complexity with respect to the number of subsequences is achieved by efficiently chunking attention into blocks with a maximum span of $H$ frames for RPE and $L$ frames for APE, as proposed in~\cite{beltagy2020longformer}. 
As a result, \modelname{}'s complexity is equivalent to that of other Transformer-based motion diffusion models~\cite{shafir2023priormdm, zhang2023diffcollage}.

\subsection{Pose-centric cross-attention}
\label{subsec:method_conditioning}
%This definition makes \modelname{} a very flexible model.
%we introduce the concept of \textit{token-exclusive conditioning}.
%Prior diffusion models for motion generations have focused on single conditioning. For example, MDM proposed to integrate the condition as an extra token~\cite{}. MotionDiffuse proposed to inject it as a scale and translation factor in the intermediate representations. However, none of these can be translated to the multi-control scenario. 

\begin{figure}

    \centering
    \includegraphics[width=\linewidth]{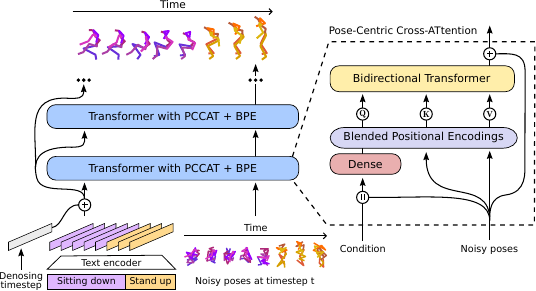}
    \caption{\textbf{Pose-centric cross-attention.} Our attention minimizes the entanglement between the control signal (e.g., text, objects) and the noisy motion by feeding the former only to the query. Consequently, our model denoises each frame's noisy pose only leveraging its own condition, and the neighboring noisy poses. }
    \vspace{-3mm}
    % Given a pose \textcolor{blue}{noisy state, and its control signal}, how can it be denoised by using only its \textcolor{orange}{neighborhood of poses}?}
    \label{fig:crossattention}

\end{figure}

In order to make motion generation with diffusion models efficient, we would like to \textit{simultaneously} generate very long sequences.
In motion Transformers, the generation is conditioned at a sequence level by injecting the condition as a token~\cite{tevet2022mdm}, or as a sequence-wise transformation in intermediate layers~\cite{zhang2022motiondiffuse}.
Therefore, they cannot be conditioned on multiple signals in different subsequences. 
For this reason, diffusion-based methods for HMC opted for individually generating sequences and then merging them~\cite{shafir2023priormdm, zhang2023diffcollage}. To enable such simultaneous heterogeneous conditioning without any extra postprocessing, we propose to inject the condition at every frame.
However, we still need to deal with a challenge: the condition never varies at training time. Therefore, at inference time, attention scores are computed with the embeddings $\embA$ and $\embB$ of the pose-condition pairs ($\poseA$, $\condA$) and ($\poseB$, $\condB$) as:
\begin{equation}\small
    q^T_mk_n=(W_q \embA)^T(W_k \embB)=\embA^T W_q^T W_k \embB.
\label{eq:entangled_attention}
\end{equation}
When $\condA {\neq} \condB$, $q^T_mk_n$ was never encountered during training. If instead of injecting the condition at every frame, we used cross-attention layers, distinct conditions would also be temporally mixed, and we would face the same problem.
To reduce the presence and impact of such training-inference misalignment, we introduce %a novel attention technique called Pose-Centric Cross-ATtention (PCCAT), see~\autoref{fig:crossattention}.
PCCAT, see~\autoref{fig:crossattention}, which aims at minimizing the entanglement between conditions and noisy poses. Specifically, PCCAT combines every frame's noisy pose and condition into queries, while using only noisy poses as keys and values. Thus, \autoref{eq:entangled_attention} becomes:
\begin{equation}\small
    \hspace{-2mm}q^T_mk_n=(W_q \embA)^T(W_k \embBnocond)=\embA^T W_q^T W_k \embBnocond.
\label{eq:pccatt_attention}
\end{equation}
With PCCAT, the attention output for pose $m$ becomes a weighted average of the value projections of its neighboring noisy poses. %, with the weights determined by its noisy pose and the conditioning information (both in the query) and its neighboring noisy pose embeddings (in the keys). 
%The Transformer output is added to the noisy poses with a residual connection.
A residual connection adds the PCCAT output to the noisy poses.
With comprehensive coverage of the motion spectrum in the training dataset, the network observes various poses preceding and following each pose, particularly within its local neighborhood. Therefore, local relationships do not suffer from unseen intermediate representations. Still, there is an obstacle to address: long-range dependencies. %, where poses from different conditional contexts need to be modeled. 
However, as discussed in \autoref{subsec:method_blendedpe}, their importance is mostly confined to the initial stages of denoising. There, the network is exposed to very noisy motion data, thus becoming robust to such unseen combinations of poses. In the latest denoising stages, when the network deals with almost clean input sequences, global dependencies have already been developed and attention is short-ranged (\autoref{fig:global_vs_local_att}).

%Therefore, out-of-distribution intermediate representations are minimized. Although this could not be the case where global dependencies are built, these are built in earlier denoising stages, where the high noise levels make the network very robust to OOD poses. 
%In order to do so, we need the motion prior to win over the conditioning signal around the transition phase.
%\textbf{Main idea:} condition is never injected to the latent representation. For that, we propose a custom self-attention where the conditions are merged only with the query. As a result, dependencies are kept simple and motion-dependent.

\begin{figure*}

    \centering
     % Figure for HumanML3D
    \includegraphics[width=\linewidth]{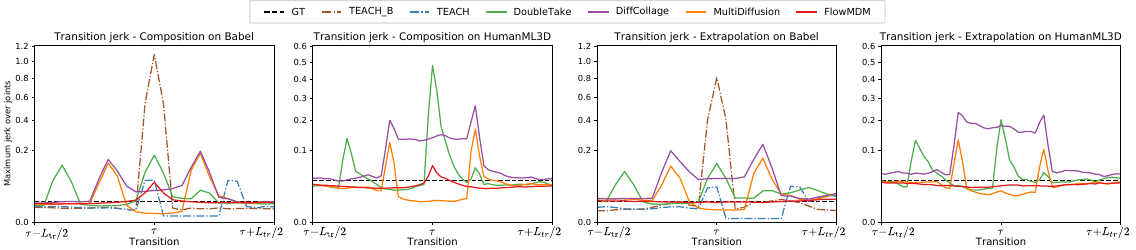}
    \vspace{-6mm}
    \caption{\textbf{Transitions smoothness.} Average maximum jerk over joints at each frame of the transitions for both motion composition (left) and extrapolation (right) tasks. While other methods show severe smoothness artifacts in the beginning and end of their transition refinement processes, \modelname{}'s jerk curve has the shortest peak for composition, and an absence of peaks for extrapolation.}
    \label{fig:jerk_plot}
    \vspace{-2mm}

\end{figure*}

\begin{table*}[t!]
\footnotesize	
\renewcommand{\arraystretch}{0}
    \centering
    %\resizebox{\linewidth}{!}{\input{tables/auto/main_comparison_babel_results}}
    \begin{tabular}{l|cccc|cccc}
\toprule
\multicolumn{1}{c}{} & \multicolumn{4}{c}{Subsequence} & \multicolumn{4}{c}{Transition} \\
  & R-prec $\uparrow{}$ & FID $\downarrow{}$ & Div $\rightarrow$ & MM-Dist $\downarrow{}$  & FID $\downarrow{}$ & Div $\rightarrow{}$ & PJ $\rightarrow{}$ & AUJ $\downarrow{}$  \\
\midrule
GT  & $0.715^{\pm0.003}$ & $0.00^{\pm0.00}$ & $8.42^{\pm0.15}$ & $3.36^{\pm0.00}$  & $0.00^{\pm0.00}$ & $6.20^{\pm0.06}$ & $0.02^{\pm0.00}$ & $0.00^{\pm0.00}$  \\
\midrule TEACH\_B  & $\textbf{0.703}^{\pm0.002}$ & $1.71^{\pm0.03}$ & $8.18^{\pm0.14}$ & $\textbf{3.43}^{\pm0.01}$  & $\underline{3.01}^{\pm0.04}$ & $\textbf{6.23}^{\pm0.05}$ & $1.09^{\pm0.00}$ & $2.35^{\pm0.01}$  \\
TEACH  & $0.655^{\pm0.002}$ & $1.82^{\pm0.02}$ & $7.96^{\pm0.11}$ & $3.72^{\pm0.01}$  & $3.27^{\pm0.04}$ & $6.14^{\pm0.06}$ & $\underline{0.07}^{\pm0.00}$ & $\underline{0.44}^{\pm0.00}$  \\
DoubleTake*  & $0.596^{\pm0.005}$ & $3.16^{\pm0.06}$ & $7.53^{\pm0.11}$ & $4.17^{\pm0.02}$  & $3.33^{\pm0.06}$ & $\underline{6.16}^{\pm0.05}$ & $0.28^{\pm0.00}$ & $1.04^{\pm0.01}$  \\
DoubleTake  & $0.668^{\pm0.005}$ & $\underline{1.33}^{\pm0.04}$ & $7.98^{\pm0.12}$ & $3.67^{\pm0.03}$  & $3.15^{\pm0.05}$ & $6.14^{\pm0.07}$ & $0.17^{\pm0.00}$ & $0.64^{\pm0.01}$  \\
MultiDiffusion  & $\underline{0.702}^{\pm0.005}$ & $1.74^{\pm0.04}$ & $\textbf{8.37}^{\pm0.13}$ & $\textbf{3.43}^{\pm0.02}$  & $6.56^{\pm0.12}$ & $5.72^{\pm0.07}$ & $0.18^{\pm0.00}$ & $0.68^{\pm0.00}$  \\
DiffCollage  & $0.671^{\pm0.003}$ & $1.45^{\pm0.05}$ & $7.93^{\pm0.09}$ & $3.71^{\pm0.01}$  & $4.36^{\pm0.09}$ & $6.09^{\pm0.08}$ & $0.19^{\pm0.00}$ & $0.84^{\pm0.01}$  \\
\midrule \modelname{}  & $\underline{0.702}^{\pm0.004}$ & $\textbf{0.99}^{\pm0.04}$ & $\underline{8.36}^{\pm0.13}$ & $3.45^{\pm0.02}$  & $\textbf{2.61}^{\pm0.06}$ & $6.47^{\pm0.05}$ & $\textbf{0.06}^{\pm0.00}$ & $\textbf{0.13}^{\pm0.00}$  \\
\bottomrule
\end{tabular}

    \vspace{-3mm}
    \caption{Comparison of \modelname{} with the state of the art in Babel. Symbols $\uparrow$, $\downarrow$, and $\rightarrow$ indicate that higher, lower, or values closer to the ground truth (GT) are better, respectively. Evaluation is run 10 times and $\pm$ specifies the 95\% confidence intervals.}
    \vspace{-1mm}
    \label{tab:babel_sota}
\end{table*}
\begin{table*}[t!]
\footnotesize	
\renewcommand{\arraystretch}{0}
    \centering
    %\resizebox{\linewidth}{!}{\input{tables/auto/main_comparison_humanml_results.tex}}
    \begin{tabular}{l|cccc|cccc}
\toprule
\multicolumn{1}{c}{} & \multicolumn{4}{c}{Subsequence} & \multicolumn{4}{c}{Transition} \\
  & R-prec $\uparrow{}$ & FID $\downarrow{}$ & Div $\rightarrow$ & MM-Dist $\downarrow{}$  & FID $\downarrow{}$ & Div $\rightarrow{}$ & PJ $\rightarrow{}$ & AUJ $\downarrow{}$  \\
\midrule
GT  & $0.796^{\pm0.004}$ & $0.00^{\pm0.00}$ & $9.34^{\pm0.08}$ & $2.97^{\pm0.01}$  & $0.00^{\pm0.00}$ & $9.54^{\pm0.15}$ & $0.04^{\pm0.00}$ & $0.07^{\pm0.00}$  \\
\midrule DoubleTake*  & $\underline{0.643}^{\pm0.005}$ & $\underline{0.80}^{\pm0.02}$ & $\underline{9.20}^{\pm0.11}$ & $\underline{3.92}^{\pm0.01}$  & $\underline{1.71}^{\pm0.05}$ & $\textbf{8.82}^{\pm0.13}$ & $0.52^{\pm0.01}$ & $2.10^{\pm0.03}$  \\
DoubleTake  & $0.628^{\pm0.005}$ & $1.25^{\pm0.04}$ & $9.09^{\pm0.12}$ & $4.01^{\pm0.01}$  & $4.19^{\pm0.09}$ & $8.45^{\pm0.09}$ & $0.48^{\pm0.00}$ & $1.83^{\pm0.02}$  \\
MultiDiffusion  & $0.629^{\pm0.002}$ & $1.19^{\pm0.03}$ & $\textbf{9.38}^{\pm0.08}$ & $4.02^{\pm0.01}$  & $4.31^{\pm0.06}$ & $8.37^{\pm0.10}$ & $\underline{0.17}^{\pm0.00}$ & $\underline{1.06}^{\pm0.01}$  \\
DiffCollage  & $0.615^{\pm0.005}$ & $1.56^{\pm0.04}$ & $8.79^{\pm0.08}$ & $4.13^{\pm0.02}$  & $4.59^{\pm0.10}$ & $8.22^{\pm0.11}$ & $0.26^{\pm0.00}$ & $2.85^{\pm0.09}$  \\
\midrule \modelname{}  & $\textbf{0.685}^{\pm0.004}$ & $\textbf{0.29}^{\pm0.01}$ & $9.58^{\pm0.12}$ & $\textbf{3.61}^{\pm0.01}$  & $\textbf{1.38}^{\pm0.05}$ & $\underline{8.79}^{\pm0.09}$ & $\textbf{0.06}^{\pm0.00}$ & $\textbf{0.51}^{\pm0.01}$  \\
\bottomrule
\end{tabular}

    \vspace{-3mm}
    \caption{Comparison of \modelname{} with the state of the art in HumanML3D.}
    \vspace{-3mm}
    \label{tab:humanml_sota}
\end{table*}

\section{Experiments}
\label{sec:results}

\def\trlength{L_{tr}}

%\textbf{Representation.} \textcolor{orange}{(right now, we keep using HumanML3D for easy implementation)} We select the SMPL thetas as our main representation, similar to \cite{azadi2023make}. We argue that the default motion representation from \cite{guo2022humanml3d} is redundant and can't be transformed to a sequence of SMPL meshes without applying an inverse kinematics optimization first, which would affect the results. The benefits are 1) direct control of an animatable avatar, and 2) coordinates information are easily computable for any body shape with forward kinematics (FK). The latter is useful when dealing with objects, where vertices positions with respect to the object are of utmost importance. The main drawback is the lack of direct control in terms of joints: a small movement in the shoulder angle represents a huge movement in the finger joints positions. \textcolor{orange}{In fact, \cite{azadi2023make} don't show animations with translation, probably because of trembling avatars.}

\subsection{Experimental setup}

\textbf{Datasets.} Our experiments are conducted on the Babel~\cite{punnakkal2021babel} and HumanML3D~\cite{guo2022humanml3d} datasets, with their train and test splits. 
HumanML3D features multiple textual descriptions of each motion sequence, but lacks explicit transition annotations, making supervised learning infeasible for transition generation. Babel, on the other hand, provides finely-grained textual descriptions at an atomic level, including transitions, which facilitates more precise and dynamic motion control but also presents a greater challenge due to fast and short transitions.
%\textbf{Motion representation.} 
To demonstrate the flexibility of \modelname{}, we employ the standard motion representations provided with each dataset. HumanML3D utilizes a 263D pose vector that includes joint coordinates, angles, velocities, and feet contact. By contrast, Babel uses the global position and orientation and a 6D rotation representation~\cite{zhou2019continuity} of the SMPL model joints~\cite{bogo2016smpl}, as in \cite{petrovich2022temos}.

\textbf{Evaluation.} Our evaluation uses the metrics established by \cite{guo2022humanml3d}, and later refined for this task in \cite{shafir2023priormdm, li2023sequential, yang2023synthesizing}. 
More specifically, motion sequences are synthesized as compositions of 32 pairs of textual descriptions and their durations. %, resulting in 32 non-overlapping subsequences $\lbrace S_i \rbrace_{1 \leq i \leq 32}$ per set. 
The 32 subsequences and the 31 transitions between $S_{i-1}$ and $S_{i}$ pairs are evaluated independently. In particular, each transition is defined as the set of consecutive poses $\lbrace{} x_{\tau_i - \trlength/2}, \dots, x_{\tau_i + \trlength/2 - 1} \rbrace{}$, sharing $\frac{\trlength}{2}$ frames with $S_{i-1}$ and $S_{i}$. The transition duration $\trlength$ is set to 30 and 60 frames for Babel and HumanML3D (1 and 3 seconds), respectively.
The top-3 R-precision (R-prec), and the multimodal distance (MM-Dist) are used to evaluate how well the subsequences' motion matches their textual description~\cite{guo2022humanml3d}. The FID score and the average pairwise distance among all motion embeddings (diversity) assess the quality and variety of both subsequences and transitions, respectively~\cite{guo2022humanml3d,heusel2017fid}. %To ensure statistical reliability, a
All metrics are averaged over 10 runs with 95\% confidence intervals reported.

\textbf{Closing the gap: the Jerk.} Generative models are hard to evaluate~\cite{sajjadi2018assessing, yang2022diffusion, creswell2018generative}. The FID score~\cite{heusel2017fid} has proven to be a very reliable metric in quantifying the similarity between distributions of generated and real motion data while being sensitive to motion artifacts or noise~\cite{maiorca2022fidevaluating}. Nevertheless, exclusively relying on perceptual metrics like FID for assessing transition quality can be misleading due to their insensitivity to motion anomalies such as abrupt accelerations~\cite{barquero2023belfusion}, or foot skating~\cite{maiorca2023fidfootskating}. To complement the FID, our work introduces two novel metrics built upon the concept of \textit{jerk} (i.e., the time derivative of acceleration), which is indicative of motion smoothness and known to be sensitive to kinetic irregularities~\cite{balasubramanian2011robust, hogan2009sensitivity, balasubramanian2015analysis, gulde2018smoothnesscomplex, larboulette2015reviewsmoothness, yi2021transpose, castillo2023bodiffusion}. Given that natural human motion typically exhibits constrained jerk due to relatively consistent acceleration patterns~\cite{larboulette2015reviewsmoothness, gulde2018smoothnesscomplex}, our metrics are tailored to highlight \textit{persistent deviations} from this norm in generated transitions. Firstly, we compute the Peak Jerk (PJ), taking the maximum value found throughout the transition motion over all joints. 
While this measure captures extreme fluctuations, it may favor models that unnaturally smooth transitions across several wider peaks of jerk.
To measure this undesirable effect, we introduce the Area Under the Jerk (AUJ), calculated as the sum of L1-norm differences between a method's instantaneous jerk and the dataset's average jerk value. This measure serves as an aggregate indicator of motion smoothness, quantifying the cumulative deviation from natural human movement across the entire transition. The PJ and AUJ of a transition are formally defined as follows:
\begin{equation}
\text{PJ} = \max_{\substack{1 \leq i \leq K \\ 1 \leq \tau \leq \trlength}} |j_i(\tau)|_{1}, \hspace{0.4cm}
\text{AUJ} = \sum_{\tau=1}^{\trlength} \max_{1 \leq i \leq K} |j_i(\tau) - j_{avg}|_{1},
\end{equation}
\noindent
where $j_i(\tau)$ is the jerk at time $\tau$ for joint $i$, $K$ is the number of joints, and $j_{avg}$ is the average joints-wise maximum jerk across the dataset.

%to use the log (\textcolor{blue}{decide if log or not}) dimensionless jerk (LDLJ),. This metric quantifies the cumulative jerk normalized by the amplitude of the motion. The jerk is the the derivative of the acceleration with respect to time. The LDLJ is computed as: \textcolor{blue}{WHAT ABOUT COMBINING XYZ?}
%\begin{align}
%j(t) &= \frac{d^3x(t)}{dt^3} & DLJ &= \frac{\int j(t)^2 \, dt}{x_m^5} \\
%LDLJ &= \log(DLJ)
%\end{align}
%The normalization is important to increase its sensitivity~\cite{hogan2009sensitivity}. In \cite{gulde2018smoothnesscomplex}, the authors proved that the LDLJ was the most appropriate metric for smoothness quantification in activities of daily living, as long as the motion durations were controlled. These two are assumptions satisfied in our scenario. One of the main disadvantages of this metric is its sensitivity to the movement duration, as longer motion accumulate more jerk over time, which is not directly correlated to less smoothness~\cite{gulde2018smoothnesscomplex}. However, since the validation motion lengths are fixed, this is not an issue. In particular, we compute the metric as the weighted sum of the LDLJ across all joints.

%\textbf{Short-to-fast transitions:} new evaluation that groups evaluated segments depending on their length. Shorter motion is more difficult to transition realistically because of its dynamic start-end of sequences.

\textbf{Baselines.} We compare our method to publicly released related works that can generate sequential motions from text: the autoregressive TEACH~\cite{athanasiou2022teach}, and the diffusion sampling techniques DoubleTake~\cite{shafir2023priormdm}, DiffCollage~\cite{zhang2023diffcollage}, and MultiDiffusion~\cite{bar2023multidiffusion}. Sampling techniques are evaluated with PCCAT and APE for a fairer comparison. 
%Since TEACH was proposed with a spherical linear interpolation to smooth transitions, we also include a baseline without it, TEACH\_B.
Additionally, we evaluate TEACH with its spherical linear interpolation over transitions turned off (TEACH\_B), and DoubleTake with MDM, as originally proposed (DoubleTake*).
%Additionally, TEACH without spherical linear interpolation over transitions, and DoubleTake with the original MDM model~\cite{tevet2022mdm} are included in the comparison as TEACH\_B, and DoubleTake*. 
TEACH and TEACH\_B cannot be trained for HumanML3D due to the lack of pairs of consecutive actions and textual descriptions.

\textbf{Implementation details.}
We tune the hyperparameters of all models with grid search. The attention horizon for RPE, $H$, is set to 100/150 for Babel/HumanML3D. %The model is trained with the Adam optimizer~\cite{}, with learning rate of X, etc. 
The number of diffusion steps is 1K for all experiments. Our model is trained with the $x_0$ parameterization~\cite{xiao2021tackling}, and minimizes the L2 reconstruction loss. During training, RPE and APE are alternated randomly at a frequency of 0.5. We use classifier-free guidance with weights 1.5/2.5~\cite{ho2022classifier}.
We use a binary step function to guide the BPE sampling, yielding 125/60 initial APE steps. 
The minimum/maximum lengths for training subsequences are set to 30/200 and 70/200 frames (i.e., 1/6.7s and 3.5/10s).
For Babel, training subsequences include consecutive ground truth motions with distinct textual descriptions in order to increase the motions variability, and make the network explicitly robust to multiple conditions. The ablation study includes two conditioning baselines: 1) concatenating each frame's condition and noisy pose, and replacing the PCCAT with vanilla self-attention (SAT), and 2) injecting the condition with cross-attention layers (CAT). See more details in \supp{} Sec. A.

\subsection{Quantitative analysis}

\begin{table*}[t!]
    \footnotesize	
    \renewcommand{\arraystretch}{0}
    \centering
    \begin{tabular}{ccc|cccc|cccc}
    
\toprule
\multicolumn{3}{c}{} & \multicolumn{4}{c}{Subsequence} & \multicolumn{4}{c}{Transition} \\
Cond. & Train. PE & Inf. PE  & R-prec $\uparrow{}$ & FID $\downarrow{}$ & Div $\rightarrow$ & MM-Dist $\downarrow{}$  & FID $\downarrow{}$ & Div $\rightarrow{}$ & PJ $\rightarrow{}$ & AUJ $\downarrow{}$  \\
\midrule
GT & - & -  & $0.715^{\pm0.003}$ & $0.00^{\pm0.00}$ & $8.42^{\pm0.15}$ & $3.36^{\pm0.00}$  & $0.00^{\pm0.00}$ & $6.20^{\pm0.06}$ & $0.02^{\pm0.00}$ & $0.00^{\pm0.00}$  \\
\midrule PCCAT & A & A  & $0.699^{\pm0.004}$ & $1.34^{\pm0.04}$ & $\textbf{8.36}^{\pm0.12}$ & $3.40^{\pm0.02}$  & $4.26^{\pm0.07}$ & $5.98^{\pm0.06}$ & $1.81^{\pm0.01}$ & $3.73^{\pm0.01}$  \\
PCCAT & R & R  & $0.635^{\pm0.006}$ & $1.28^{\pm0.03}$ & $8.05^{\pm0.11}$ & $4.02^{\pm0.02}$  & $2.18^{\pm0.07}$ & $\textbf{6.14}^{\pm0.08}$ & $\underline{0.03}^{\pm0.00}$ & $0.20^{\pm0.00}$  \\
\midrule PCCAT & B & A  & $\underline{0.716}^{\pm0.006}$ & $1.20^{\pm0.04}$ & $8.31^{\pm0.14}$ & $\underline{3.32}^{\pm0.02}$  & $3.01^{\pm0.06}$ & $6.35^{\pm0.07}$ & $1.78^{\pm0.01}$ & $3.66^{\pm0.02}$  \\
PCCAT & B & R  & $0.635^{\pm0.004}$ & $\textbf{0.85}^{\pm0.02}$ & $8.25^{\pm0.12}$ & $3.98^{\pm0.02}$  & $\underline{2.14}^{\pm0.04}$ & $6.44^{\pm0.09}$ & $0.04^{\pm0.00}$ & $0.15^{\pm0.00}$  \\
\midrule SAT & B & B  & $0.681^{\pm0.004}$ & $1.52^{\pm0.04}$ & $8.22^{\pm0.11}$ & $3.61^{\pm0.02}$  & $\textbf{1.91}^{\pm0.03}$ & $6.41^{\pm0.07}$ & $0.06^{\pm0.00}$ & $\underline{0.12}^{\pm0.00}$  \\
CAT & B & B  & $\textbf{0.719}^{\pm0.004}$ & $1.29^{\pm0.02}$ & $8.16^{\pm0.13}$ & $\textbf{3.27}^{\pm0.02}$  & $2.57^{\pm0.08}$ & $\underline{6.06}^{\pm0.07}$ & $\textbf{0.02}^{\pm0.00}$ & $\textbf{0.07}^{\pm0.00}$  \\
\midrule PCCAT & B & B  & $0.702^{\pm0.004}$ & $\underline{0.99}^{\pm0.04}$ & $\textbf{8.36}^{\pm0.13}$ & $3.45^{\pm0.02}$  & $2.61^{\pm0.06}$ & $6.47^{\pm0.05}$ & $0.06^{\pm0.00}$ & $0.13^{\pm0.00}$  \\
\bottomrule
\end{tabular}

    %\resizebox{\linewidth}{!}{\input{tables/auto/ablation_babel_results.tex}}
    \vspace{-3mm}
    \caption{Ablation study in Babel. Cond. indicates the conditioning scheme, Train./Inf. PE specify the positional encodings (PE) used at training/inference time, and A, R, and B refer to absolute, relative, and blended PE, respectively. $\uparrow$, $\downarrow$, and $\rightarrow$ indicate that higher, lower, or values closer to the ground truth (GT) are better, respectively. Evaluation is run 10 times and $\pm$ specifies the 95\% confidence intervals.}
    \label{tab:babel_ablation}
    \vspace{-1mm}
\end{table*}

\begin{table*}[t!]
    \footnotesize	
    \renewcommand{\arraystretch}{0}
    \centering
    \begin{tabular}{ccc|cccc|cccc}
\toprule
\multicolumn{3}{c}{} & \multicolumn{4}{c}{Subsequence} & \multicolumn{4}{c}{Transition} \\
Cond. & Train. PE & Inf. PE  & R-prec $\uparrow{}$ & FID $\downarrow{}$ & Div $\rightarrow$ & MM-Dist $\downarrow{}$  & FID $\downarrow{}$ & Div $\rightarrow{}$ & PJ $\rightarrow{}$ & AUJ $\downarrow{}$  \\
\midrule
GT & - & -  & $0.796^{\pm0.004}$ & $0.00^{\pm0.00}$ & $9.34^{\pm0.08}$ & $2.97^{\pm0.01}$  & $0.00^{\pm0.00}$ & $9.54^{\pm0.15}$ & $0.04^{\pm0.00}$ & $0.07^{\pm0.00}$  \\
\midrule PCCAT & A & A  & $0.689^{\pm0.005}$ & $0.66^{\pm0.02}$ & $9.73^{\pm0.12}$ & $3.63^{\pm0.02}$  & $3.90^{\pm0.12}$ & $8.29^{\pm0.08}$ & $1.50^{\pm0.01}$ & $3.40^{\pm0.02}$  \\
PCCAT & R & R  & $0.531^{\pm0.005}$ & $1.75^{\pm0.07}$ & $8.71^{\pm0.10}$ & $4.80^{\pm0.03}$  & $2.53^{\pm0.12}$ & $8.62^{\pm0.08}$ & $\textbf{0.03}^{\pm0.00}$ & $0.58^{\pm0.01}$  \\
\midrule PCCAT & B & A  & $\textbf{0.699}^{\pm0.005}$ & $0.61^{\pm0.02}$ & $9.76^{\pm0.10}$ & $\underline{3.54}^{\pm0.02}$  & $2.42^{\pm0.09}$ & $8.39^{\pm0.09}$ & $1.40^{\pm0.01}$ & $3.29^{\pm0.02}$  \\
PCCAT & B & R  & $0.554^{\pm0.007}$ & $1.06^{\pm0.06}$ & $9.02^{\pm0.11}$ & $4.54^{\pm0.02}$  & $\textbf{1.12}^{\pm0.04}$ & $\textbf{9.00}^{\pm0.10}$ & $0.05^{\pm0.00}$ & $0.53^{\pm0.01}$  \\
\midrule SAT & B & B  & $\underline{0.692}^{\pm0.004}$ & $\underline{0.49}^{\pm0.02}$ & $\underline{9.08}^{\pm0.09}$ & $\textbf{3.51}^{\pm0.01}$  & $3.19^{\pm0.08}$ & $8.09^{\pm0.11}$ & $\underline{0.04}^{\pm0.00}$ & $\textbf{0.36}^{\pm0.02}$  \\
CAT & B & B  & $0.622^{\pm0.005}$ & $1.27^{\pm0.04}$ & $8.86^{\pm0.15}$ & $4.10^{\pm0.01}$  & $3.93^{\pm0.14}$ & $8.23^{\pm0.10}$ & $\underline{0.04}^{\pm0.00}$ & $\underline{0.49}^{\pm0.02}$  \\
\midrule PCCAT & B & B  & $0.685^{\pm0.004}$ & $\textbf{0.29}^{\pm0.01}$ & $\textbf{9.58}^{\pm0.12}$ & $3.61^{\pm0.01}$  & $\underline{1.38}^{\pm0.05}$ & $\underline{8.79}^{\pm0.09}$ & $0.06^{\pm0.00}$ & $0.51^{\pm0.01}$  \\
\bottomrule
\end{tabular}

    %\resizebox{\linewidth}{!}{\input{tables/auto/ablation_humanml_results.tex}}
    \vspace{-3mm}
    \caption{Ablation study in HumanML3D.}
    \label{tab:humanml_ablation}
    \vspace{-3mm}
\end{table*}

\textbf{Comparison with the state of the art on HMC.} 
Tables \ref{tab:babel_sota} and \ref{tab:humanml_sota} show the comparison of \modelname{} with current state-of-the-art models in Babel and HumanML3D datasets, respectively. In HumanML3D, our model outperforms by a fair margin the other methods in terms of subsequence accuracy-wise metrics (R-prec and MM-Dist), and FID. In Babel, it matches the state of the art in accuracy and excels in FID score. \modelname{} produces transitions of higher quality and smoothness on both datasets, as indicated by FID, PJ, and AUJ metrics.
The lack of correlation between the FID score and the AUJ underscores the importance of the latter as a complementary metric for assessing smoothness. \autoref{fig:jerk_plot}-left shows the average jerk values across the generated transitions. We observe that state-of-the-art methods exhibit severe smoothness artifacts. During TEACH's spherical linear interpolation, the jerk quickly reaches values near zero. By contrast, DiffCollage leans toward higher-than-average jerk values, while MultiDiffusion exhibits the opposite trend. DoubleTake shows three peaks, caused by their two-stage noise estimation process. In comparison, \modelname{} successfully minimizes peak jerk values, producing the smoothest transitions between subsequences. See \supp{} Sec. C for in-depth analyses.

%\subsection{Human Motion Extrapolation}
\textbf{Human motion extrapolation.} In single text-to-motion, the duration of the generated motion is limited to the longest subsequence length $L$ available in the training set. Extrapolating periodic actions into sequences longer than those in the ground truth presents a notable challenge. Achieving this through HMC requires the harmonization of periodicity across adjacent subsequences. However, common strategies that combine independently generated subsequences often disrupt the periodicity of the motion.
To assess our model's capabilities in addressing this issue, we construct an evaluation set comprising 32 consecutive repetitions of 32 different extrapolatable actions such as `walk forward', `jumping', or `playing the guitar', extracted from the Babel and HumanML3D test sets (more details in \supp{} Sec. B).
\autoref{fig:jerk_plot}-right displays the motion jerk across transitions for all models on this task. We observe that, while other models exhibit smoothness anomalies similar to those shown in the HMC evaluation, \modelname{} closely mirrors the ground truth jerk. This observation indicates that the jerk peak noted in \modelname{} for the composition task is likely attributed to smoothness irregularities in more complex transitions.

\begin{figure}[t!]

    \centering
    \includegraphics[width=\linewidth]{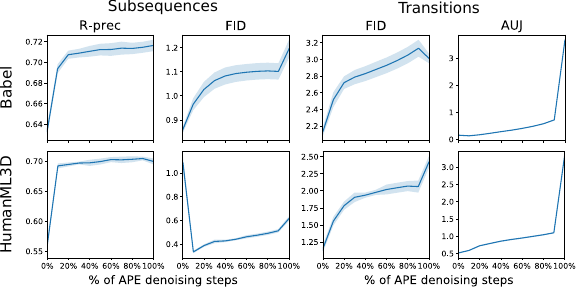}
    \caption{\textbf{BPE trade-offs.} Increasing the number of APE steps undergone during BPE sampling improves the correspondence between motion and textual description (R-prec), but reduces the transition realism and smoothness (FID and AUJ). The best balance is reached around 10\% of APE denoising steps.}
    \vspace{-5mm}
    \label{fig:inference_schedule}

\end{figure}
%\begin{figure}
%    \centering
%    \begin{minipage}[b]{0.48\linewidth}
%        \includegraphics[width=\linewidth]{figures/extrapolation_humanml_transition_Jerk.pdf}
%    \end{minipage}
%    \hfill % This will insert a non-breaking space between the two minipages
    % Figure for Babel
%    \begin{minipage}[b]{0.48\linewidth}
%        \includegraphics[width=\linewidth]{figures/extrapolation_babel_transition_Jerk.pdf}
%    \end{minipage}
%    \caption{\textbf{Extrapolation smoothness.} Plots showing the extrapolation capabilities of RPE/BPE in terms of smoothness. \modelname{} resembles the smoothness levels of the ground truth. \textcolor{red}{(Mayb combine with Fig. 4??. + ADD DoubleTake + REMOVE APE/RPE)}}
%    \label{fig:extrapolation}

%\end{figure}

%IDEA: show plot for transition flow in different extrapolation scenarios against the SOTA (e.g., walk, run, walk in circles, throw something, etc.)
%A common problem on sequence modeling is how to train with sequences within a size $[0,S)$, and generate sequences longer than $S$. This is called an extrapolation problem. While prior works have attempted to extrapolate well with specific relative positional embeddings, we show that our framework extrapolates very well. In fact, while having only seen a person walking in circles for at most X seconds, \modelname{} is able to generate a person walking in circles for an unlimited amount of time with no degeneration.

% ========================

% ========================

\begin{figure*}[t!]
    \centering
    \includegraphics[width=\linewidth]{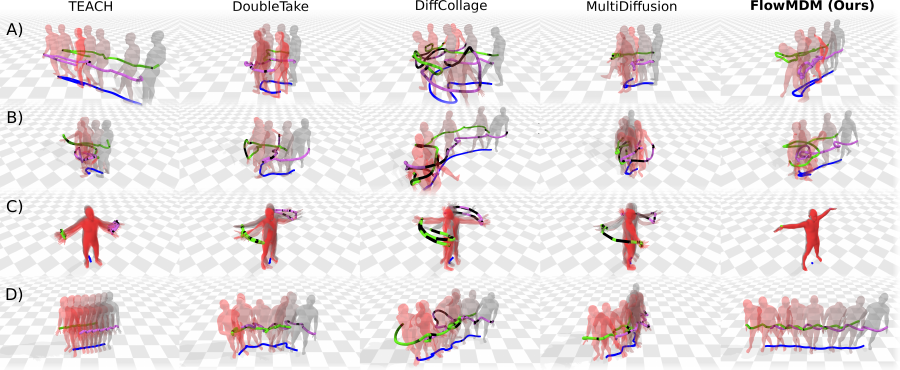}
    \vspace{-6mm}
    \caption{\textbf{Qualitative analysis (Babel).} A) and B) show compositions of 3 motions (`walk straight'$\xrightarrow{}$`side steps'$\xrightarrow{}$`walk backward', and `walk'$\xrightarrow{}$`turn around'$\xrightarrow{}$`sit on the bench', respectively), and C) and D) illustrate extrapolations that repeat 6 times a static (`t-pose') and a dynamic (`step to the right') action, respectively. Solid curves match the trajectories of the global position (blue) and left/right hands (purple/green). Darker colors indicate instantaneous jerk deviations from the median value, saturating at twice the jerk's standard deviation in the dataset (black segments). %, with a saturation point at twice the standard deviation of the dataset. 
    Abrupt transitions manifest as black segments amidst lighter ones. 
    \modelname{} exhibits the most fluid motion and preserves the staticity or periodicity of extrapolated actions, in contrast to other methods that show spontaneous high jerk values and fail to keep the motion coherence in extrapolations.}
    \label{fig:exp_qualitative}
    \vspace{-4mm}
\end{figure*}

\textbf{Ablation study.} 
The effectiveness of BPE and PCCAT is presented in Tables~\ref{tab:babel_ablation} and \ref{tab:humanml_ablation}. Reasonably, the baseline model trained solely with APE fails to generate smooth transitions. Conversely, a model trained only with RPE, despite producing the smoothest transitions, struggles to model global motion dependencies and accurately reflect the corresponding textual descriptions. 
Interestingly, training with BPE improves the performance of both APE- and RPE-only samplings.
%The results also show that training with BPE is not detrimental, as both aforementioned behaviors can be replicated by sampling from the same model with either APE or RPE. 
Sampling with BPE combines the best of both worlds by preserving the excellent AUJ values of the RPE models and reaching the state-of-the-art accuracy and FID scores of the APE models. \autoref{fig:inference_schedule} illustrates this balance. Specifically, increasing the number of APE steps enhances the motion's congruence with the textual description, at the cost of reducing the smoothness and realism of the transitions.
In HumanML3D, the SAT and CAT conditioning schemes lead to worse transitions in terms of FID and diversity. This is caused by the coexistence of different conditions in the local neighborhood of the transition at inference, which never happens during training. 
Our PCCAT conditioning technique effectively solves this problem. In Babel, such effect is not present because the training motion sequences include several subsequences, thus increasing the model's robustness to transitions with varying conditions.

%\textbf{GOAL 1.} BPE allows to generate smooth and realistic transitions while keeping the global intra-sequence coherence.

%\textbf{GOAL 2.} PCCAT performs great when transitions supervision is nonexistent. In fully unsupervised scenarios like HumanML3D, OOD transitions are critical and make SAT and CAT conditioning strategies degenerate.

%\textbf{In- vs. out-of-distribution sequences.} Babel validation set is an example.
%\textbf{Short-, medium-, and long-action composition.} HumanML3D three splits. OR go to \supp{}.

%For example, PriorMDM uses a refinement window of 50-60 frames, which would make the generation of 60-frames long consecutive segments impossible (i.e., 2 seconds). Our method is able to handle these cases correctly. We argue that generating short segments in a dataset as dynamic as Babel is important.

%\subsection{Synthetic data is all you need}
%Infer a whole new synthetic dataset, provide a description, and train for other underrepresented downstream tasks. Examples of these:
%\begin{itemize}
%    \item GOAL: from initial pose, reach an object.
%    \item HOI: from initial pose/motion, predict interaction with an object,
%\end{itemize}

\textbf{On the efficiency of \modelname{}.}
%In \ref{fig:nfe_analysis}, we show the NFE of \modelname{} compared to competitors. 
Diffusion-based state-of-the-art methods such as MultiDiffusion and DiffCollage denoise poses from the transition more than once in order to harmonize it with the adjacent motions. DoubleTake's transitions undergo an additional denoising process, which adds computational burden and can not be parallelized. Oppositely, \modelname{} does not apply redundant denoising steps to any pose. In particular, our model goes through 47.1\%, 28.4\%, and 16.5\% less pose-wise denoising steps than DoubleTake, DiffCollage, and MultiDiffusion%MultiDiffusion, DiffCollage, and DoubleTake
, respectively.

\subsection{Qualitative results}

\autoref{fig:exp_qualitative} illustrates how our quantitative findings translate into visual outcomes on the human motion composition and extrapolation tasks. First, as anticipated by \autoref{fig:jerk_plot}, we confirm that state-of-the-art methods produce short intervals of jerk peaks around transitions. These do not typically match long-range motion scenarios, where such jerks might be contextually appropriate. Contrarily, \modelname{} produces motion that is realistic, accurate, and smooth. Particularly, we notice that DiffCollage's bias toward producing constantly high jerk values around transitions is perceived as an overall chaotic motion. Due to the independent generation of their subsequences, DoubleTake, DiffCollage, and MultiDiffusion are unable to maintain the static or periodic nature of actions when extrapolating them. Only TEACH and \modelname{} are able to successfully extrapolate a static `t-pose', and ours is the only one capable of extrapolating a 'step to the right' sequence realistically. Finally, \modelname{} also inherits the trajectory control capabilities of motion diffusion models as shown in \autoref{fig:intro}-right.

%In terms of global coherence, TEACH and \modelname{} are the only ones that can explicitly model sequences of consecutive actions, so they show the best coherence in terms of combined trajectory.
%DoubleTake, DiffCollage, and MultiDiffusion generate subsequences independently, which compromises the global motion coherence. By contrast, \modelname{}'s relative denoising stage is trained with consecutive sequences, which makes it able to 
\section{Conclusion}
\label{sec:conclusions}

We presented \modelname{}, the first approach that generates human motion compositions simultaneously, without undergoing postprocessing or redundant denoising diffusion steps. We also %showed the advantages of relative position encodings in motion modeling, and 
introduced the blended positional encodings to combine the benefits of absolute and relative positional encodings during the denoising chain. Finally, we presented the pose-centric cross-attention, a technique that improves the generation of transitions when training with only a single condition per motion sequence.
%training with only a single textual description per motion sequence.

\textbf{Limitations and future work.} The absolute stage of BPE does not model relationships between subsequences. Consequently, their low-frequency spectrum is generated independently. This limitation could be addressed in future work by incorporating an intention planning module. 
Finally, our method learns a strong motion prior that generates transitions between combinations of actions never seen at training time. Such capability 
could theoretically be used with different models leveraging different control signals, assuming they all are trained under the same framework. Future work will experimentally validate this hypothesis. 
%, which is not handled by \modelname{}.

%\textbf{Limitations.} While the simplicity of our approach is one of its main advantages, our transition modeling is based on a learnt prior whose robustness depends on the particular scenario and the diversity of motions in the dataset used for training.
%While the quadratic complexity of attention could make our approach slower than its competitors, methods to linearise the Transformer complexity are available. Thanks to the encoder-only architecture of our method, such speed-ups are applicable (with iterative decoding, the main bottleneck is not the quadratic attention).

%%%%%%%%% REFERENCES
{\small
\bibliographystyle{ieee_fullname}
\bibliography{egbib}
}

%%%%%%%%%% SUPPLEMENTARY
\clearpage
\noindent {\Large \textbf{Supplementary Material}\par}
\vspace{0.5cm}

\renewcommand*{\thesection}{\Alph{section}}
\renewcommand*{\thefigure}{\Alph{figure}}
\renewcommand*{\thetable}{\Alph{table}}
\setcounter{section}{0}
\setcounter{table}{0}
\setcounter{figure}{0}
\section{Further implementation details}
\label{sec:supp_implementation_details}

All values are reported as X/Y for Babel/HumanML3D, or as Z if values are equal for both. Note that motion sequences are downsampled to 30/20 fps.

\textbf{State-of-the-art models.} TEACH is used off-the-shelf~\footnote{\url{https://github.com/athn-nik/teach/commit/f4285aff0fd556a5b46518a751fc90825d91e68b}} with the originally proposed alignment and spherical linear interpolation, and without them (TEACH\_B). DoubleTake is used off-the-shelf~\footnote{\url{https://github.com/priorMDM/priorMDM/commit/8bc565b3120c08182f067e161e83403b0efe7cc9}} from their original repository, with the parameters \textit{handshake size} and \textit{blending length} set to 10/20f (frames), and 10/5f, respectively. To fulfill the constraints of their method, the handshake size needs to be shorter than half the shortest sequence we want to generate, which is 30f (1s) for Babel. Since DoubleTake uses the original Motion Diffusion Model~\cite{tevet2022mdm}, whose training discarded very short sequences, it underperforms in our more comprehensive evaluation protocol (see \autoref{sec:supp_eval_details}).
For a fairer comparison, we also evaluate it using our diffusion model with absolute positional encodings (APE), and call it DoubleTake*. DoubleTake* uses the same handshake size and blending length as DoubleTake. DiffCollage and MultiDiffusion were implemented manually, and utilize our model as well for the same reasons mentioned earlier. We set their sampling parameter \textit{transition length} to 10/20f. For DoubleTake, DiffCollage, and MultiDiffusion, we use classifier-free guidance with weights 1.5/2.5 during sampling.

\textbf{\modelname{}.} Our diffusion model uses 1k steps and a cosine noise schedule~\cite{nichol2021improved}. \modelname{} is trained with the $x_0$ parameterization~\cite{xiao2021tackling}, and an L2 reconstruction loss. Denoising timesteps are encoded as a sinusoidal positional encoding that goes through two dense layers into a 512D vector. Textual descriptions are tokenized and embedded with CLIP~\cite{radford2021learning} into 512D vectors. Poses of 135/263D are encoded by a dense layer into a sequence of 512D vectors. If the APE is active, a sinusoidal encoding is added to the embedded poses at this stage. Then, the embedded poses are taken as the \textit{keys} and \textit{values} of a Transformer. Embedded poses are concatenated to the sum of the timesteps and text embeddings, and fed to a dense layer. The resulting 512D vectors are the \textit{queries}. If the relative positional encoding (RPE) is active, rotary embeddings~\cite{su2021roformer} are injected to the queries and keys at this stage. The output of the Transformer is added to the embedded poses with a residual connection. 8~Transformers are stacked together. A final dense layer converts the pose embeddings back to a vector of 135/263D, which are the denoised poses. A dropout of 0.1 is applied to the APE, and to the inputs of the Transformers. The attention span of the Transformers is capped within each subsequence during the APE stage, and within the attention horizon H=100/150f during the RPE stage. We train with blended positional encodings (BPE), i.e.,  RPE and APE are alternated randomly at a frequency of 0.5. We use Adam~\cite{kingma2014adam} with learning rate of 0.0001 as our optimizer, and train for 1.3M/500k steps in a single RTX 3090 (about 4/2 days).
During BPE sampling, the binary step schedule transitions from absolute to relative mode after 125/60 denoising steps (out of 1k steps). Classifier-free guidance with weights 1.5/2.5 is used during sampling.

\section{Evaluation details}
\label{sec:supp_eval_details}

Generative models are difficult to evaluate and compare due to the limitations of the metrics (discussed in Sec.~4.1) and the stochasticity present during sampling. To alleviate the latter, we run all our evaluation 10 times and provide the 95\% confidence intervals. However, we still face another issue in our task: the randomness in the combinations of textual descriptions. The generation difficulty for the combination `sit down'$\rightarrow$`stand up'$\rightarrow$`run' is not the same as for `sit down'$\rightarrow$`run'$\rightarrow$`stand up'. The evaluation protocol from \cite{shafir2023priormdm} includes 32 evaluation sequences of 32 randomly sampled textual descriptions from the test set. The generated motion needs to perform sequentially the 32 actions from each evaluation sequence. However, these descriptions are sampled differently in each evaluation run, which hinders reproducibility. In order to ensure proper replication and a fair comparison in future works, we propose a more thorough and fully reproducible evaluation protocol that enables a more fine-grained analysis based on \textit{scenarios} (analysis provided in \autoref{subsec:supp_scenario_wise}):

\textbf{Babel.} We built two scenarios with in-distribution (50\%) and out-of-distribution (50\%) combinations. For the in-distribution scenario, we first selected test motion sequences showcasing at least three consecutive actions (i.e., textual descriptions) with a total duration of 1.5s. Then, we randomly sampled from them to build 32 sets of 32 combinations of textual descriptions.
%32 sequences are sampled so that it contains from the consecutive textual descriptions present in the test set. 
For the out-of-distribution scenario, 32 sets were built by autoregressively sampling 32 textual descriptions so that consecutive actions did not appear together neither in the training nor in the test set.

\textbf{HumanML3D.} Since annotations in HumanML3D do not include consecutive actions, we cannot build in- and out-of-distribution scenarios. However, this dataset contains a great variability of sequence lengths (3-10s). Therefore, we decided to build four scenarios by varying the length of the subsequences included. More specifically, we created three sets of 6, 8, and 18 combinations (9.4, 12.5, 28.1\%) by sampling 32 short (3-5s), medium (5-8s), and long (8-10s) test motions, respectively. Ratios were set so that all together preserved the proportion of short, medium, and long subsequences in the original test set. This is important to keep the validity of statistical measures like FID. Additionally, we included another scenario with 32 sets (50\%) of 32 random motion sequences from the test set.

We share the list of evaluation combinations for both the human motion composition and extrapolation tasks in our public code repository\footnote{\url{https://barquerogerman.github.io/FlowMDM/}}. Note that a combination consists of a list of textual descriptions and their associated durations. The 32 textual descriptions used for the extrapolation experiments from Sec.~4 are enumerated in~\autoref{tab:extrapolated_sequences}.

%\subsection{Building the combinations}
%\subsection{Extrapolation experiments}

{\renewcommand{\arraystretch}{2.2}\linespread{0.5}\setlength{\tabcolsep}{1pt}
\begin{table}[h]
    \footnotesize
    \centering
    \begin{tabular}{p{0.42\linewidth} p{0.56\linewidth}}
        \toprule
        Babel & HumanML3D \\
        \midrule
        walk forward & a person walks in a curved path to the left. \\
        swim movement & a person stands still and does not move. \\
        stretch arms & a person walks straight forward. \\
        walk & a person does jumping jacks. \\
        stand & a person start to dance with legs. \\
        step backwards & person walking in an s shape. \\
        t-pose & a person walks to his right. \\
        throw the ball & a person slowly walked forward. \\
        run & the person is standing still doing body stretches. \\
        circle right arm backwards & the person is dancing the waltz. \\
        wave right & the person is clapping. \\
        ginga dance & walking side to side. \\
        forward kick & a person stayed on the place. \\
        look around & person is jogging in place. \\
        steps to the right & a person walks backward for 3 steps. \\
        side steps & person is running in a circle. \\
        hop forward & the person is waving hi. \\
        dance with arms & a person walks in a circular path. \\
        jog & swinging arms up and down. \\
        walk slowly & a man walks counterclockwise in a circle. \\
        jump jacks series & the person is walking towards the left. \\
        run in half a circle & the person is walking on the treadmill. \\
        walk a few steps ahead & the man is moving his left arm. \\
        move head up and down & the person is doing basketball signals. \\
        rotate right ankle & a person remained sitting down. \\
        play guitar & a person hits his drums. \\
        jump forward & person is doing a dance. \\
        move both hands around chest & a person takes some steps forward. \\
        swing back and forth & a person slowly walks forward five steps. \\
        wave & a person jumps in place. \\
        shake it & this person appears to be painting. \\
        walk in circle & a person wiping a surface with something. \\
        \bottomrule
    \end{tabular}
    \vspace{-2mm}
    \caption{Extrapolated motions for Babel and HumanML3D.}
    \label{tab:extrapolated_sequences}
\end{table}
}

\section{More experimental results}
\label{sec:supp_experimental_results}

%=========================================================

\subsection{Fine-grained comparison}
\label{subsec:supp_scenario_wise}

\begin{table*}[]
    \centering
    \footnotesize
    \begin{tabular}{l|cccc|cccc}
\toprule
\multicolumn{1}{c}{} & \multicolumn{4}{c}{Subsequence} & \multicolumn{4}{c}{Transition} \\
  & R-prec $\uparrow{}$ & FID $\downarrow{}$ & Div $\rightarrow$ & MM-Dist $\downarrow{}$  & FID $\downarrow{}$ & Div $\rightarrow{}$ & PJ $\rightarrow{}$ & AUJ $\downarrow{}$  \\
\midrule
GT  & $0.715^{\pm0.003}$ & $0.00^{\pm0.00}$ & $8.42^{\pm0.15}$ & $3.36^{\pm0.00}$  & $0.00^{\pm0.00}$ & $6.20^{\pm0.06}$ & $0.02^{\pm0.00}$ & $0.00^{\pm0.00}$  \\

\midrule In-distribution\\
\midrule TEACH\_B  & $\textbf{0.727}^{\pm0.004}$ & $2.26^{\pm0.03}$ & $8.20^{\pm0.12}$ & $\textbf{3.35}^{\pm0.01}$  & $\underline{2.77}^{\pm0.05}$ & $6.32^{\pm0.07}$ & $1.03^{\pm0.00}$ & $2.20^{\pm0.01}$  \\
TEACH  & $0.665^{\pm0.003}$ & $2.09^{\pm0.03}$ & $8.06^{\pm0.09}$ & $3.73^{\pm0.02}$  & $2.78^{\pm0.06}$ & $6.31^{\pm0.07}$ & $\underline{0.07}^{\pm0.00}$ & $\underline{0.42}^{\pm0.01}$  \\
DoubleTake*  & $0.620^{\pm0.006}$ & $3.04^{\pm0.06}$ & $7.49^{\pm0.07}$ & $4.19^{\pm0.02}$  & $3.04^{\pm0.12}$ & $\underline{6.21}^{\pm0.06}$ & $0.28^{\pm0.00}$ & $1.01^{\pm0.01}$  \\
DoubleTake  & $0.682^{\pm0.008}$ & $\underline{1.52}^{\pm0.03}$ & $7.90^{\pm0.07}$ & $3.67^{\pm0.04}$  & $3.47^{\pm0.08}$ & $6.16^{\pm0.07}$ & $0.17^{\pm0.00}$ & $0.62^{\pm0.01}$  \\
MultiDiffusion  & $0.724^{\pm0.008}$ & $2.00^{\pm0.05}$ & $\underline{8.36}^{\pm0.10}$ & $\underline{3.38}^{\pm0.02}$  & $6.33^{\pm0.13}$ & $5.91^{\pm0.06}$ & $0.17^{\pm0.00}$ & $0.65^{\pm0.01}$  \\
DiffCollage  & $0.690^{\pm0.006}$ & $1.92^{\pm0.07}$ & $7.92^{\pm0.09}$ & $3.67^{\pm0.02}$  & $4.25^{\pm0.15}$ & $\textbf{6.19}^{\pm0.07}$ & $0.19^{\pm0.01}$ & $0.82^{\pm0.02}$  \\
\modelname{} (Ours)  & $\underline{0.726}^{\pm0.006}$ & $\textbf{1.36}^{\pm0.05}$ & $\textbf{8.47}^{\pm0.10}$ & $3.40^{\pm0.03}$  & $\textbf{2.26}^{\pm0.08}$ & $6.60^{\pm0.08}$ & $\textbf{0.05}^{\pm0.00}$ & $\textbf{0.11}^{\pm0.00}$  \\

\midrule Out-of-distribution\\
\midrule TEACH\_B  & $\underline{0.680}^{\pm0.006}$ & $1.75^{\pm0.04}$ & $8.15^{\pm0.11}$ & $3.51^{\pm0.01}$  & $3.53^{\pm0.06}$ & $\underline{6.04}^{\pm0.10}$ & $1.14^{\pm0.01}$ & $2.49^{\pm0.01}$  \\
TEACH  & $0.644^{\pm0.004}$ & $2.06^{\pm0.03}$ & $7.94^{\pm0.12}$ & $3.70^{\pm0.01}$  & $4.08^{\pm0.08}$ & $6.00^{\pm0.09}$ & $\textbf{0.07}^{\pm0.00}$ & $\underline{0.46}^{\pm0.00}$  \\
DoubleTake*  & $0.572^{\pm0.007}$ & $3.78^{\pm0.07}$ & $7.53^{\pm0.12}$ & $4.15^{\pm0.02}$  & $3.83^{\pm0.09}$ & $\textbf{6.12}^{\pm0.07}$ & $0.28^{\pm0.00}$ & $1.07^{\pm0.02}$  \\
DoubleTake  & $0.654^{\pm0.009}$ & $1.65^{\pm0.07}$ & $8.06^{\pm0.08}$ & $3.66^{\pm0.02}$  & $\textbf{2.98}^{\pm0.06}$ & $6.03^{\pm0.07}$ & $0.17^{\pm0.00}$ & $0.66^{\pm0.01}$  \\
MultiDiffusion  & $\textbf{0.681}^{\pm0.009}$ & $2.11^{\pm0.06}$ & $\textbf{8.35}^{\pm0.08}$ & $\textbf{3.47}^{\pm0.03}$  & $6.97^{\pm0.12}$ & $5.67^{\pm0.05}$ & $0.19^{\pm0.00}$ & $0.71^{\pm0.01}$  \\
DiffCollage  & $0.652^{\pm0.004}$ & $\underline{1.60}^{\pm0.07}$ & $7.91^{\pm0.09}$ & $3.74^{\pm0.01}$  & $4.65^{\pm0.19}$ & $6.00^{\pm0.09}$ & $0.20^{\pm0.00}$ & $0.86^{\pm0.01}$  \\
\modelname{} (Ours)  & $0.679^{\pm0.004}$ & $\textbf{1.26}^{\pm0.06}$ & $\underline{8.16}^{\pm0.08}$ & $\underline{3.50}^{\pm0.03}$  & $\underline{3.17}^{\pm0.12}$ & $6.44^{\pm0.09}$ & $\textbf{0.07}^{\pm0.00}$ & $\textbf{0.17}^{\pm0.00}$  \\
\bottomrule
\end{tabular}

    \vspace{-3mm}
    \caption{Scenario-wise comparison in Babel. Symbols $\uparrow$, $\downarrow$, and $\rightarrow$ indicate that higher, lower, or values closer to the ground truth (GT) are better, respectively. Evaluation is run 10 times and $\pm$ specifies the 95\% confidence intervals.}
    \label{tab:babel_scenarios}
\end{table*}

\begin{table*}[]
    \centering
    \footnotesize
    \begin{tabular}{l|cccc|cccc}
\toprule
\multicolumn{1}{c}{} & \multicolumn{4}{c}{Subsequence} & \multicolumn{4}{c}{Transition} \\
  & R-prec $\uparrow{}$ & FID $\downarrow{}$ & Div $\rightarrow$ & MM-Dist $\downarrow{}$  & FID $\downarrow{}$ & Div $\rightarrow{}$ & PJ $\rightarrow{}$ & AUJ $\downarrow{}$  \\
\midrule
GT  & $0.796^{\pm0.004}$ & $0.00^{\pm0.00}$ & $9.34^{\pm0.08}$ & $2.97^{\pm0.01}$  & $0.00^{\pm0.00}$ & $9.54^{\pm0.15}$ & $0.04^{\pm0.00}$ & $0.07^{\pm0.00}$  \\
\midrule Short \\
\midrule DoubleTake*  & $0.649^{\pm0.012}$ & $\textbf{3.03}^{\pm0.18}$ & $\textbf{9.52}^{\pm0.11}$ & $3.72^{\pm0.05}$  & $\underline{3.56}^{\pm0.14}$ & $\underline{8.92}^{\pm0.14}$ & $0.13^{\pm0.01}$ & $0.79^{\pm0.05}$  \\
DoubleTake  & $0.704^{\pm0.022}$ & $4.85^{\pm0.20}$ & $10.01^{\pm0.15}$ & $\underline{3.25}^{\pm0.09}$  & $4.40^{\pm0.24}$ & $8.88^{\pm0.17}$ & $\underline{0.09}^{\pm0.00}$ & $\underline{0.73}^{\pm0.02}$  \\
MultiDiffusion  & $\textbf{0.717}^{\pm0.011}$ & $5.49^{\pm0.15}$ & $10.14^{\pm0.17}$ & $\textbf{3.23}^{\pm0.07}$  & $4.66^{\pm0.27}$ & $8.68^{\pm0.08}$ & $0.10^{\pm0.00}$ & $0.92^{\pm0.02}$  \\
DiffCollage  & $0.705^{\pm0.012}$ & $\underline{4.69}^{\pm0.18}$ & $\underline{9.73}^{\pm0.14}$ & $3.30^{\pm0.04}$  & $4.81^{\pm0.32}$ & $8.49^{\pm0.12}$ & $0.15^{\pm0.01}$ & $1.13^{\pm0.10}$  \\
\modelname{} (Ours)  & $\underline{0.714}^{\pm0.015}$ & $4.75^{\pm0.26}$ & $9.90^{\pm0.20}$ & $3.31^{\pm0.06}$  & $\textbf{3.17}^{\pm0.17}$ & $\textbf{9.03}^{\pm0.14}$ & $\textbf{0.04}^{\pm0.00}$ & $\textbf{0.59}^{\pm0.04}$  \\

\midrule Medium \\
\midrule DoubleTake*  & $0.644^{\pm0.009}$ & $\underline{2.18}^{\pm0.08}$ & $\underline{9.18}^{\pm0.12}$ & $3.72^{\pm0.04}$  & $\textbf{3.34}^{\pm0.30}$ & $\textbf{8.73}^{\pm0.12}$ & $0.14^{\pm0.00}$ & $\underline{0.70}^{\pm0.03}$  \\
DoubleTake  & $0.642^{\pm0.014}$ & $2.34^{\pm0.05}$ & $9.59^{\pm0.09}$ & $3.79^{\pm0.05}$  & $5.42^{\pm0.30}$ & $\underline{8.61}^{\pm0.11}$ & $0.12^{\pm0.00}$ & $0.83^{\pm0.02}$  \\
MultiDiffusion  & $\textbf{0.673}^{\pm0.007}$ & $3.22^{\pm0.10}$ & $9.91^{\pm0.07}$ & $\textbf{3.54}^{\pm0.04}$  & $6.24^{\pm0.34}$ & $8.11^{\pm0.12}$ & $\underline{0.10}^{\pm0.00}$ & $1.14^{\pm0.01}$  \\
DiffCollage  & $0.661^{\pm0.010}$ & $\textbf{2.03}^{\pm0.07}$ & $\textbf{9.38}^{\pm0.10}$ & $3.60^{\pm0.04}$  & $4.95^{\pm0.27}$ & $8.13^{\pm0.09}$ & $0.14^{\pm0.00}$ & $\textbf{0.66}^{\pm0.05}$  \\
\modelname{} (Ours)  & $\underline{0.669}^{\pm0.012}$ & $3.18^{\pm0.15}$ & $9.68^{\pm0.08}$ & $\underline{3.55}^{\pm0.04}$  & $\underline{4.18}^{\pm0.43}$ & $8.52^{\pm0.07}$ & $\textbf{0.04}^{\pm0.00}$ & $0.86^{\pm0.03}$  \\

\midrule Long \\
\midrule DoubleTake*  & $\underline{0.616}^{\pm0.006}$ & $\underline{2.51}^{\pm0.09}$ & $\underline{8.77}^{\pm0.08}$ & $\underline{4.09}^{\pm0.03}$  & $\underline{3.38}^{\pm0.18}$ & $\underline{8.50}^{\pm0.11}$ & $0.89^{\pm0.02}$ & $3.52^{\pm0.07}$  \\
DoubleTake  & $0.605^{\pm0.006}$ & $4.07^{\pm0.13}$ & $8.19^{\pm0.11}$ & $4.18^{\pm0.01}$  & $8.45^{\pm0.33}$ & $7.79^{\pm0.12}$ & $0.81^{\pm0.02}$ & $3.04^{\pm0.07}$  \\
MultiDiffusion  & $0.569^{\pm0.012}$ & $5.02^{\pm0.15}$ & $8.07^{\pm0.07}$ & $4.49^{\pm0.05}$  & $8.56^{\pm0.32}$ & $7.91^{\pm0.10}$ & $\underline{0.23}^{\pm0.01}$ & $\underline{1.16}^{\pm0.01}$  \\
DiffCollage  & $0.557^{\pm0.008}$ & $5.79^{\pm0.13}$ & $7.75^{\pm0.09}$ & $4.61^{\pm0.02}$  & $9.00^{\pm0.36}$ & $7.75^{\pm0.09}$ & $0.38^{\pm0.01}$ & $5.04^{\pm0.14}$  \\
\modelname{} (Ours)  & $\textbf{0.666}^{\pm0.012}$ & $\textbf{1.93}^{\pm0.08}$ & $\textbf{8.81}^{\pm0.09}$ & $\textbf{3.81}^{\pm0.04}$  & $\textbf{2.85}^{\pm0.22}$ & $\textbf{8.54}^{\pm0.11}$ & $\textbf{0.08}^{\pm0.00}$ & $\textbf{0.45}^{\pm0.03}$  \\

\midrule All \\
\midrule DoubleTake*  & $\underline{0.655}^{\pm0.007}$ & $\underline{0.84}^{\pm0.04}$ & $\textbf{9.29}^{\pm0.10}$ & $\underline{3.92}^{\pm0.03}$  & $\underline{1.91}^{\pm0.12}$ & $\textbf{8.79}^{\pm0.11}$ & $0.51^{\pm0.01}$ & $2.11^{\pm0.05}$  \\
DoubleTake  & $0.621^{\pm0.006}$ & $1.49^{\pm0.07}$ & $8.91^{\pm0.04}$ & $4.13^{\pm0.02}$  & $4.75^{\pm0.13}$ & $8.39^{\pm0.06}$ & $0.47^{\pm0.01}$ & $1.84^{\pm0.03}$  \\
MultiDiffusion  & $0.632^{\pm0.003}$ & $1.17^{\pm0.04}$ & $\textbf{9.29}^{\pm0.09}$ & $4.05^{\pm0.02}$  & $4.42^{\pm0.16}$ & $8.37^{\pm0.08}$ & $\underline{0.17}^{\pm0.00}$ & $\underline{1.06}^{\pm0.01}$  \\
DiffCollage  & $0.615^{\pm0.007}$ & $1.73^{\pm0.07}$ & $8.73^{\pm0.05}$ & $4.18^{\pm0.04}$  & $4.98^{\pm0.24}$ & $8.09^{\pm0.06}$ & $0.26^{\pm0.00}$ & $2.71^{\pm0.12}$  \\
\modelname{}  & $\textbf{0.695}^{\pm0.008}$ & $\textbf{0.30}^{\pm0.02}$ & $9.55^{\pm0.08}$ & $\textbf{3.58}^{\pm0.02}$  & $\textbf{1.49}^{\pm0.06}$ & $\underline{8.78}^{\pm0.11}$ & $\textbf{0.06}^{\pm0.00}$ & $\textbf{0.50}^{\pm0.01}$  \\
\bottomrule
\end{tabular}

    \vspace{-3mm}
    \caption{Scenario-wise comparison in HumanML3D.}
    \label{tab:humanml_scenarios}
\end{table*}

\autoref{tab:babel_scenarios} shows the comparison of \modelname{} with the state of the art in both in-distribution and out-of-distribution scenarios. We observe that, while all methods maintain similar performance in both scenarios for the subsequence generation, they generate less realistic and more abrupt transitions in the out-of-distribution case. \modelname{} performs the best at most metrics in both scenarios, with an important gap with respect to the previous state of the art regarding transition smoothness. \autoref{tab:humanml_scenarios} shows the scenario-wise results for HumanML3D, where \modelname{} also performs the best in most metrics and scenarios. Interestingly, MultiDiffusion is, after ours, the most stable method in terms of transition smoothness across scenarios (PJ and AUJ), whereas DiffCollage and DoubleTake show severe transition degeneration in combinations of long sequences. Such degeneration is mostly due to their methodological need to pad the motion sequence during sampling. When dealing with long sequences, sequences might be extended beyond the maximum sequence length at training time. Therefore, given that the APE does not extrapolate well, the generation in the padded motion, or transition, tends to degenerate. Our method naturally avoids this limitation.

%=========================================================

\subsection{On the attention horizon}
\label{subsec:supp_attention_scores}

\begin{table*}[]
    \centering
    \footnotesize
    \begin{tabular}{cc|cccc|cccc}
\toprule
\multicolumn{2}{c}{} & \multicolumn{4}{c}{Subsequence} & \multicolumn{4}{c}{Transition} \\
 H (frames)& Inf. PE   & R-prec $\uparrow{}$ & FID $\downarrow{}$ & Div $\rightarrow$ & MM-Dist $\downarrow{}$  & FID $\downarrow{}$ & Div $\rightarrow{}$ & PJ $\rightarrow{}$ & AUJ $\downarrow{}$  \\
\midrule
GT & -  & $0.715^{\pm0.003}$ & $0.00^{\pm0.00}$ & $8.42^{\pm0.15}$ & $3.36^{\pm0.00}$  & $0.00^{\pm0.00}$ & $6.20^{\pm0.06}$ & $0.02^{\pm0.00}$ & $0.00^{\pm0.00}$  \\
\midrule 50 & R  & $0.641^{\pm0.004}$ & $1.03^{\pm0.04}$ & $7.99^{\pm0.11}$ & $3.92^{\pm0.03}$  & $\textbf{2.04}^{\pm0.06}$ & $6.30^{\pm0.05}$ & $\textbf{0.04}^{\pm0.00}$ & $0.15^{\pm0.00}$  \\
100 & R  & $0.635^{\pm0.004}$ & $\textbf{0.85}^{\pm0.02}$ & $8.25^{\pm0.12}$ & $3.98^{\pm0.02}$  & $\underline{2.14}^{\pm0.04}$ & $6.44^{\pm0.09}$ & $\textbf{0.04}^{\pm0.00}$ & $0.15^{\pm0.00}$  \\
150 & R  & $0.641^{\pm0.005}$ & $\underline{0.99}^{\pm0.04}$ & $8.24^{\pm0.15}$ & $3.88^{\pm0.03}$  & $2.43^{\pm0.06}$ & $6.43^{\pm0.06}$ & $\textbf{0.04}^{\pm0.00}$ & $0.15^{\pm0.00}$  \\
200 & R  & $0.601^{\pm0.005}$ & $1.48^{\pm0.04}$ & $7.85^{\pm0.14}$ & $4.17^{\pm0.02}$  & $3.18^{\pm0.09}$ & $\textbf{6.16}^{\pm0.05}$ & $\textbf{0.04}^{\pm0.00}$ & $0.19^{\pm0.00}$  \\
\midrule 50 & B  & $0.698^{\pm0.006}$ & $1.07^{\pm0.03}$ & $8.19^{\pm0.11}$ & $3.44^{\pm0.02}$  & $2.34^{\pm0.06}$ & $\underline{6.24}^{\pm0.07}$ & $0.06^{\pm0.00}$ & $\textbf{0.13}^{\pm0.00}$  \\
100 & B  & $\underline{0.702}^{\pm0.004}$ & $\underline{0.99}^{\pm0.04}$ & $\textbf{8.36}^{\pm0.13}$ & $3.45^{\pm0.02}$  & $2.61^{\pm0.06}$ & $6.47^{\pm0.05}$ & $0.06^{\pm0.00}$ & $\textbf{0.13}^{\pm0.00}$  \\
150 & B  & $\textbf{0.704}^{\pm0.004}$ & $1.24^{\pm0.03}$ & $\underline{8.34}^{\pm0.12}$ & $\underline{3.43}^{\pm0.02}$  & $2.54^{\pm0.08}$ & $6.40^{\pm0.08}$ & $0.06^{\pm0.00}$ & $\textbf{0.13}^{\pm0.00}$  \\
200 & B  & $0.694^{\pm0.006}$ & $1.13^{\pm0.02}$ & $8.25^{\pm0.13}$ & $\textbf{3.42}^{\pm0.02}$  & $3.31^{\pm0.08}$ & $6.38^{\pm0.09}$ & $0.06^{\pm0.00}$ & $0.14^{\pm0.01}$  \\
\bottomrule
\end{tabular}
    \vspace{-3mm}
    \caption{Attention horizon effect in Babel. All models correspond to \modelname{}, trained with BPE. Inf. PE indicates the type of positional encoding used during sampling: B for BPE, and R for only RPE. Symbols $\uparrow$, $\downarrow$, and $\rightarrow$ indicate that higher, lower, or values closer to the ground truth (GT) are better, respectively. Evaluation is run 10 times and $\pm$ specifies the 95\% confidence intervals.}
    \label{tab:babel_atthorizon}
\end{table*}

\begin{table*}[]
    \centering
    \footnotesize
    \begin{tabular}{cc|cccc|cccc}
\toprule
\multicolumn{2}{c}{} & \multicolumn{4}{c}{Subsequence} & \multicolumn{4}{c}{Transition} \\
H (frames) & Inf. PE   & R-prec $\uparrow{}$ & FID $\downarrow{}$ & Div $\rightarrow$ & MM-Dist $\downarrow{}$  & FID $\downarrow{}$ & Div $\rightarrow{}$ & PJ $\rightarrow{}$ & AUJ $\downarrow{}$  \\
\midrule
GT & -  & $0.796^{\pm0.004}$ & $0.00^{\pm0.00}$ & $9.34^{\pm0.08}$ & $2.97^{\pm0.01}$  & $0.00^{\pm0.00}$ & $9.54^{\pm0.15}$ & $0.04^{\pm0.00}$ & $0.07^{\pm0.00}$  \\
\midrule 50 & R  & $0.583^{\pm0.005}$ & $1.08^{\pm0.07}$ & $9.03^{\pm0.15}$ & $4.30^{\pm0.02}$  & $1.88^{\pm0.06}$ & $8.85^{\pm0.10}$ & $\underline{0.04}^{\pm0.00}$ & $0.70^{\pm0.01}$  \\
100 & R  & $0.591^{\pm0.005}$ & $1.07^{\pm0.03}$ & $9.02^{\pm0.13}$ & $4.29^{\pm0.02}$  & $1.51^{\pm0.08}$ & $8.90^{\pm0.08}$ & $\underline{0.04}^{\pm0.00}$ & $0.56^{\pm0.01}$  \\
150 & R  & $0.554^{\pm0.007}$ & $1.06^{\pm0.06}$ & $9.02^{\pm0.11}$ & $4.54^{\pm0.02}$  & $\textbf{1.12}^{\pm0.04}$ & $\textbf{9.00}^{\pm0.10}$ & $0.05^{\pm0.00}$ & $0.53^{\pm0.01}$  \\
200 & R  & $0.528^{\pm0.004}$ & $1.37^{\pm0.04}$ & $8.87^{\pm0.07}$ & $4.68^{\pm0.01}$  & $1.72^{\pm0.05}$ & $\underline{8.97}^{\pm0.09}$ & $\textbf{0.03}^{\pm0.00}$ & $0.97^{\pm0.01}$  \\
\midrule 50 & B  & $0.671^{\pm0.004}$ & $\textbf{0.25}^{\pm0.01}$ & $\textbf{9.37}^{\pm0.14}$ & $3.66^{\pm0.02}$  & $\underline{1.27}^{\pm0.04}$ & $8.79^{\pm0.08}$ & $0.06^{\pm0.00}$ & $\underline{0.52}^{\pm0.01}$  \\
100 & B  & $\underline{0.684}^{\pm0.003}$ & $0.36^{\pm0.02}$ & $9.55^{\pm0.09}$ & $\textbf{3.61}^{\pm0.02}$  & $2.04^{\pm0.11}$ & $8.59^{\pm0.06}$ & $0.06^{\pm0.00}$ & $0.56^{\pm0.01}$  \\
150 & B  & $\textbf{0.685}^{\pm0.004}$ & $\underline{0.29}^{\pm0.01}$ & $9.58^{\pm0.12}$ & $\textbf{3.61}^{\pm0.01}$  & $1.38^{\pm0.05}$ & $8.79^{\pm0.09}$ & $0.06^{\pm0.00}$ & $\textbf{0.51}^{\pm0.01}$  \\
200 & B  & $0.658^{\pm0.006}$ & $0.47^{\pm0.03}$ & $\textbf{9.37}^{\pm0.13}$ & $3.77^{\pm0.02}$  & $2.27^{\pm0.07}$ & $8.69^{\pm0.08}$ & $0.06^{\pm0.00}$ & $0.68^{\pm0.01}$  \\
\bottomrule
\end{tabular}

    \vspace{-3mm}
    \caption{Attention horizon effect in HumanML3D. All models correspond to \modelname{}, trained with BPE. Inf. PE indicates the type of positional encoding used during sampling: B for BPE, and R for only RPE.}
    \label{tab:humanml_atthorizon}
\end{table*}

%Fig.~2 from the main paper showed how the attention during later denoising stages mainly focuses on the local neighborhood of the pose to be denoised.
In Tabs. \ref{tab:babel_atthorizon} and \ref{tab:humanml_atthorizon}, we show the effect of the attention horizon when using RPE for either a purely relative inference schedule, or our proposed BPE inference schedule. We observe how increasing it too much (H=200) makes the network perform worse at transition generation in both datasets (FID and AUJ), and also in subsequence generation for HumanML3D (R-prec and MM-Dist). Conversely, when decreasing it too much (H=50), the capacity to model long-range dynamics becomes limited, thus reducing the accuracy of the generated subsequences (R-prec and MM-Dist). As the performance with H of 100 and 150 is similar in both datasets, we chose values that are closest to the average sequence length in each dataset, i.e., 100/150f for Babel/HumanML3D.

%For \modelname{}, we selected the H value that maximizes the smoothness metrics (AUJ) and keep a high subsequence quality and accuracy (FID and R-prec). These are 100/150 for Babel/HumanML3D.
% The best values, 100/150f for Babel/HumanML3D, correspond to the average length of the motion sequences seen at training time.

%=========================================================

\subsection{On the diffusion schedule}
\label{subsec:supp_diffusion_schedule}

\begin{figure}
    \centering
    \includegraphics[width=\linewidth]{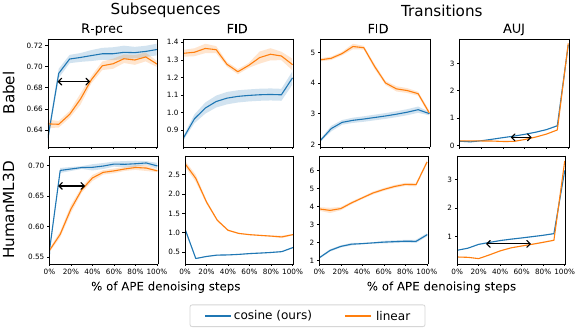}
    \vspace{-6mm}
    \caption{\textbf{Diffusion noise schedules.} The cosine noise schedule destroys the motion signal slower and in a more evenly distributed way than the linear schedule. As a result, \modelname{} is able to exploit better the low-to-high frequencies decomposition along the denoising chain and generate better subsequences and transitions. The faster motion destruction in the linear schedule translates to needing more APE steps to reconstruct global dependencies inside subsequences (black arrows $\leftrightarrow$).}
    \label{fig:supp_schedules_noise}
\end{figure}

The discussion and the BPE design in Sec.~3.2 are motivated by the low-to-high frequencies decomposition during the denoising stage of diffusion models. However, the denoising process depends on how the noise is injected, or the \textit{noise schedule}. The linear and the cosine (our choice) noise schedules are the most common schedules. The linear schedule destroys the motion very fast, reaching a non-recognizable state after going through the 75\% of the diffusion steps~\cite{nichol2021improved}. Instead, the cosine schedule destroys the motion signal slower and in a more evenly distributed way. \autoref{fig:supp_schedules_noise} shows the performance of \modelname{} during BPE sampling with both schedules. First, we observe that \modelname{} benefits from the steadier noise injection of the cosine schedule, achieving better performance in all realism and accuracy metrics (R-prec and FID). Second, we identify a displacement in the accuracy (R-prec) and smoothness (AUJ) curves (see black arrows). Given that with the linear schedule global dependencies start being recovered later, more APE steps are needed to achieve the accuracy and smoothness reached with the cosine schedule.

%=========================================================

%\subsection{On the condition role}
%\label{subsec:supp_condition_importance}

%It would be interesting to show the strength of the motion prior in later denoising stages. There is a tradeoff between condition and motion prior guidances. We can check the effect of dropping the condition in later denoising stages, and the difference between scores (y-axis: $\mid f(x_t, c)-f(x_t, \emptyset)\mid$, x-axis: $t$). Mention that future work could explore schedules for the guidance score along the denoising chain.

%Also related, show the attention scores of the condition across the CAT stages.

%=========================================================
\subsection{On the classifier-free guidance}
\label{subsec:supp_cfg}

\begin{figure}
    \centering
    \includegraphics[width=\linewidth]{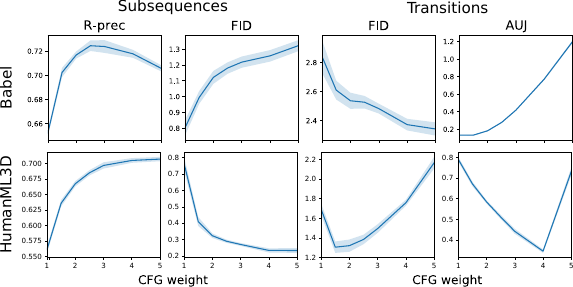}
    \vspace{-6mm}
    \caption{\textbf{Classifier-free guidance. }In line with prior works, we also observe an accuracy improvement (R-prec) when increasing the strength (i.e., \textit{weight}) of the classifier-free guidance (CFG). However, above certain values, the performance degrades, especially in terms of smoothness (AUJ). This is caused by the misalignment of CFG directions on each side of the transition.}
    \label{fig:cfg_effects}
\end{figure}

The classifier-free guidance is an important add-on for diffusion sampling that intensifies the conditioning signal, thus improving the quality and accuracy of the generated samples~\cite{ho2022classifier}. It is implemented by first computing the conditionally denoised motion $x_c$, and the unconditionally denoised motion $x$. Then, the denoised sample is computed as $x+w(x_c-x)$. If $w{=}1$, the classifier-free guidance is deactivated. When generating motion from single textual descriptions with classifier-free guidance, we keep steering the denoising toward motions matching better the textual description. However, when building human motion compositions with our method, two different conditions coexist in the neighborhoods of the transitions. There, the classifier-free guidance pushes the denoising towards dispar directions. As a result, if $w$ is too high, the transition will become sharper, and if $w$ is too low, subsequences might not be accurate enough. \autoref{fig:cfg_effects} shows these effects for \modelname{}. We notice a sweet point around $w{=}1.5/2.5$ for Babel/HumanML3D, where \modelname{} reaches the maximum accuracy and quality for subsequences and a good trade-off for quality and smoothness of transitions.
\section{Qualitative results}
\label{subsec:supp_qualitative}

\begin{figure*}
    \centering
    \includegraphics[width=\linewidth]{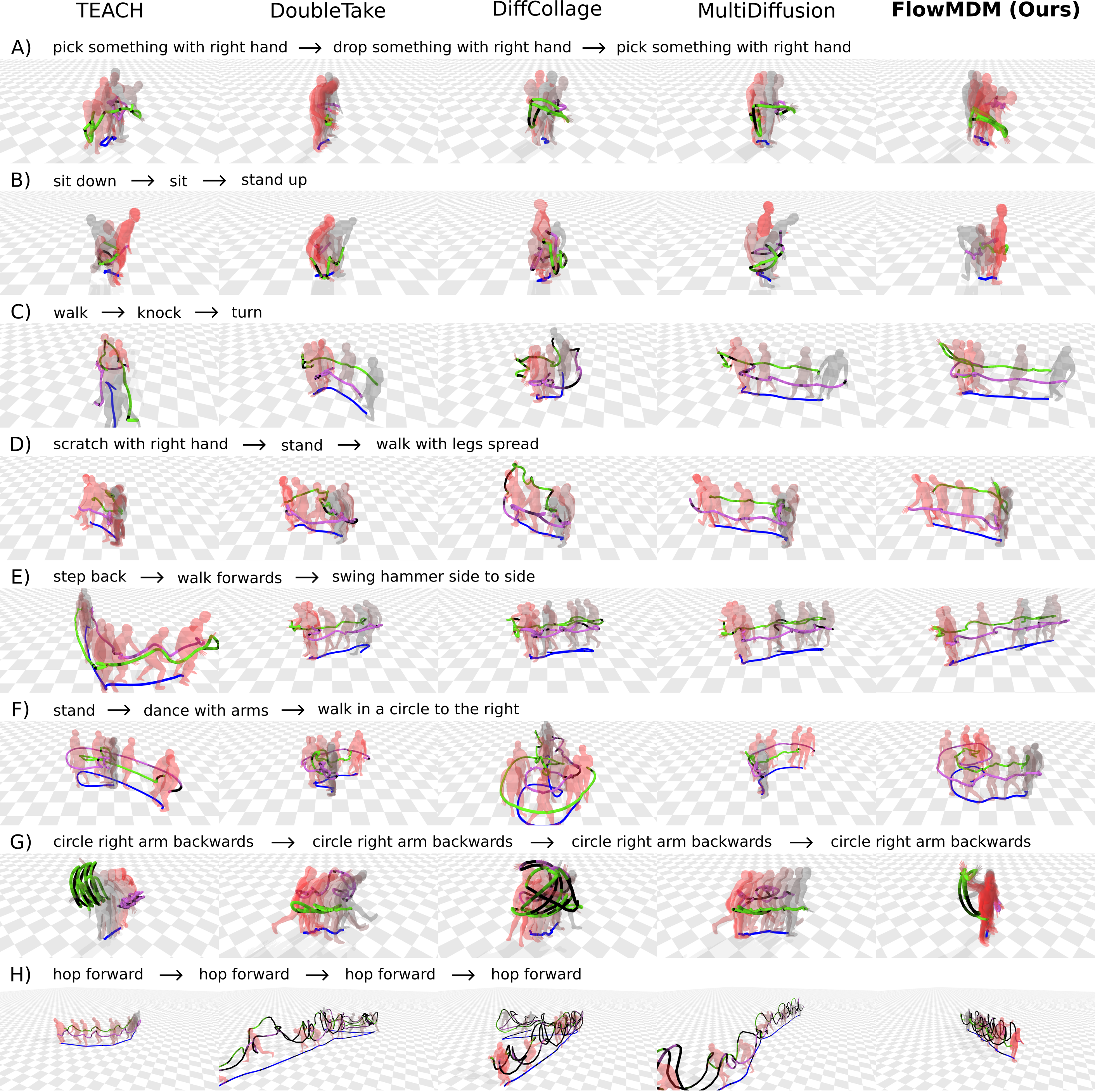}
    \vspace{-6mm}
    \caption{\textbf{Qualitative examples (Babel).} A-F feature six human motion compositions, and G-H two human motion extrapolations. According to the scenarios defined in \autoref{sec:supp_eval_details}, A, B, C belong to in-distribution combinations, and D, E, F to out-of-distribution combinations. Videos of all samples are also included as part of this supplementary material. Solid curves match the trajectories of the global position (blue) and left/right hands (purple/green). Darker colors indicate instantaneous jerk deviations from the median value, saturating at twice the jerk’s standard deviation in the dataset (black segments). Abrupt transitions manifest as black segments amidst lighter ones.}
    \label{fig:supp_babel_qualitative}
\end{figure*}

\begin{figure*}
    \centering
    \includegraphics[width=\linewidth]{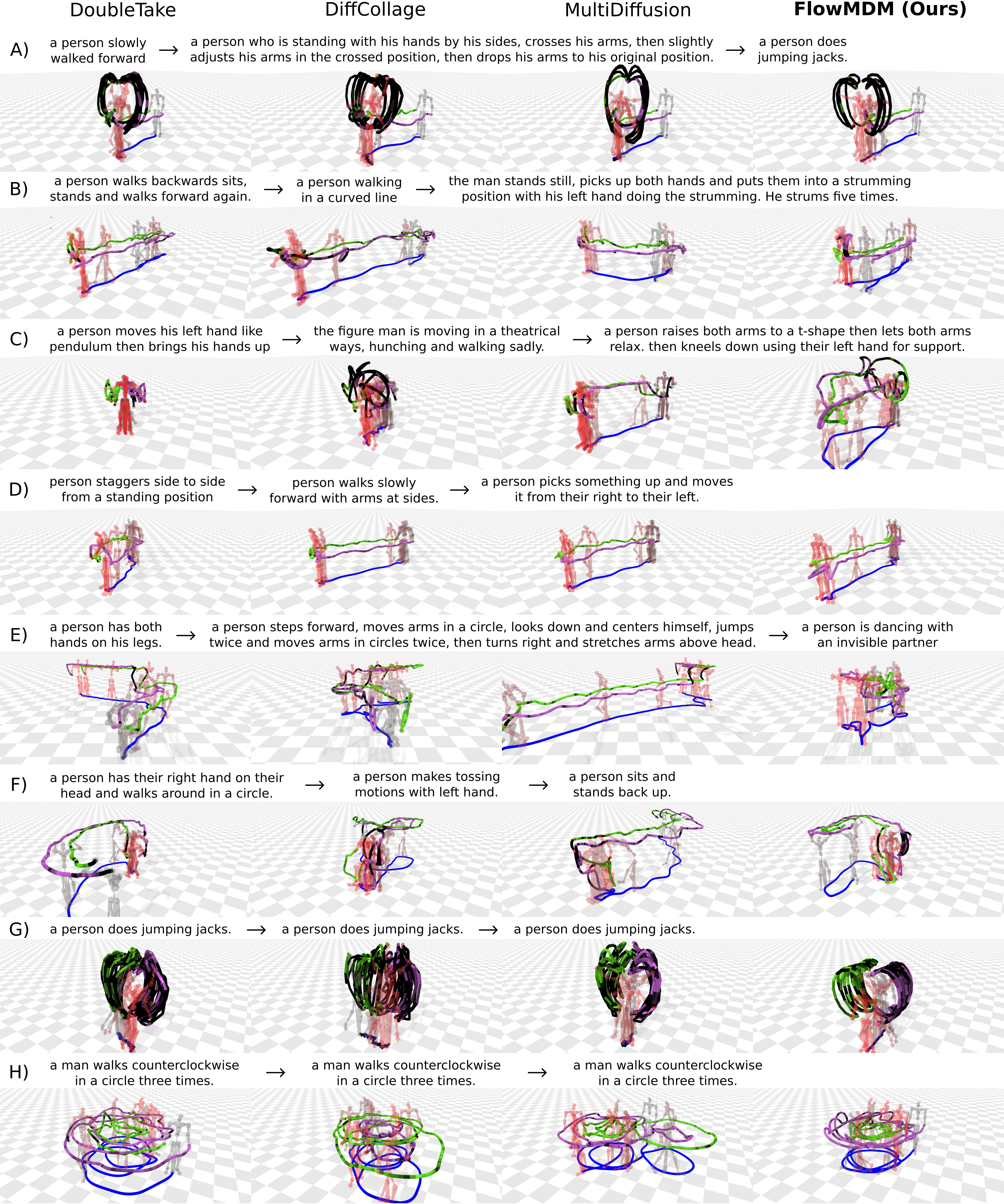}
    \vspace{-6mm}
    \caption{\textbf{Qualitative examples (HumanML3D).} A-F feature six human motion compositions, and G-H two human motion extrapolations. According to the scenarios defined in \autoref{sec:supp_eval_details}, A, B, C are samples from the short, medium, and long scenarios, respectively, and D, E, F from the mixed scenario. Videos of all samples are also included as part of this supplementary material.}
    \label{fig:supp_humanml_qualitative}
\end{figure*}

Figs. \ref{fig:supp_babel_qualitative} and \ref{fig:supp_humanml_qualitative} show six human motion compositions (A to F), and two extrapolations (G and H) for Babel and HumanML3D, respectively. The compositions are subsets of the evaluation combinations composed of 32 actions, so the beginning and end of these can contain partial transitions toward other actions. Motion videos are also included as part of the supplementary material. Note that we can represent the motions from Babel with SMPL body meshes thanks to its motion representation including the SMPL parameters~\cite{bogo2016smpl}. For HumanML3D, we use skeletons, as its motion representation only includes the 3D coordinates of the joints.

\textbf{Discussion. } 
The hands trajectories and the jerk color indicators in Figs.~\ref{fig:supp_babel_qualitative} and \ref{fig:supp_humanml_qualitative} and the videos highlight that \modelname{} generates the smoothest transitions between subsequences. Notably, state-of-the-art methods exhibit frequent smoothness artifacts (black segments) in the boundaries of their transitions. We notice that the compositions produced by TEACH lack realism due to the use of a naive spherical linear interpolation, disrupting the motion dynamics. This becomes more apparent in extrapolations G and H of both datasets, where the periodicity of the movement is clearly compromised. 
On the other side, DoubleTake, DiffCollage, and MultiDiffusion share two significant limitations. Firstly, they adhere to a predetermined transition length, which may not fit all situations. For example, in Babel-A, the `picking' actions occur very rapidly due to the insufficient length for generating a natural transition. By contrast, our approach is able to leverage more transitioning time from either transition side if needed,
%is able to take the time needed from any side of the transition, 
without artificial constraints. Secondly, the denoising process in these methods only considers a small portion of the neighboring subsequences, leading to poor performance in dynamic motion extrapolations. For example, in HumanML3D-G, they all generate erratic jumping jacks. While our method also independently generates the low-frequency motion spectrum, it effectively rectifies inconsistencies in later stages, yielding realistic and periodic motion. In the case of Babel-H, where successfully extrapolating the `hop forward' action requires synchronizing each subsequence with the whole neighboring motion, our model is the only one able to generate a smooth, coherent, and realistic extrapolation.

\textbf{Limitations. } 
However, \modelname{} is not without its imperfections. We noticed that our method struggles with very complex descriptions, such as the first one in HumanML3D-B. Instead of executing the intricate description that includes `walk backwards, sit, stand, and walk forward again', it only walks backwards. Given that the partial execution of actions is also observed in other methods, we consider it a challenge associated with the broader text-to-motion task.  Indeed, our model could theoretically also benefit from improved conditioning schemes such as using better text embeddings.
%potential loss in semantics when embedding such complex textual descriptions with CLIP~\cite{nichol2021improved}. 
Another acknowledged limitation of our model, discussed in Sec.~5, is the independent generation of low-frequency components. In Babel-B, for example, a slight mismatch between the sitting and standing positions is observed.
Nonetheless, in contrast to DiffCollage, MultiDiffusion, and DoubleTake which also exhibit this effect, \modelname{} produces a smoother result.

\end{document}